\documentclass{ieeeaccess}
\usepackage{xcolor}
\usepackage{cite}
\usepackage{amsmath,amssymb,amsfonts}
\usepackage{algorithmic}
\usepackage{textcomp}
\usepackage{float}
\usepackage{soul}
\usepackage{changepage}
\usepackage{graphicx}
\usepackage{subcaption}
\usepackage{multirow}
\usepackage{comment}

\definecolor{blue}{RGB}{0,0,255}
\usepackage{CJKutf8}
\def\BibTeX{{\rm B\kern-.05em{\sc i\kern-.025em b}\kern-.08em
T\kern-.1667em\lower.7ex\hbox{E}\kern-.125emX}}
\usepackage{pict2e}
\newsavebox{\ORCIDlogo}
\savebox{\ORCIDlogo}{%
\setlength{\unitlength}{\dimexpr 1em/256\relax}%
\begin{picture}(256,256)%
  \color[HTML]{A6CE39}\put(128,128){\circle*{256}}%
  \color{white}%
  \put(78.6,199.2){\circle*{20}}%
  \moveto(70.9,176,9)\lineto(86.3,176,9)\lineto(86.3,69.8)\lineto(70.9,69.8)%
  \closepath\fillpath%
  \moveto(108.9,176.9)\lineto(150.5,176.9)%
  \curveto(190.1,176.9)(207.5,148.6)(207.5 ,123.3)%
  \curveto(207.5,95,8)(186,69.7)(150.7,69.7)%
  \lineto(108.9,69.7)%
  \closepath\fillpath%
  \color[HTML]{A6CE39}%
  \moveto(124.3,83.6)\lineto(148.8,83.6)%
  \curveto(183.7,83.6)(191.7,110.1)(191.7,123.3)%
  \curveto(191.7,144.8)(178,163)(148,163)%
  \lineto(124.3,163)%
  \closepath\fillpath%
\end{picture}%
}
\newcommand\orcidicon[1]{\href{https://orcid.org/#1}{\usebox{\ORCIDlogo}}}
\usepackage{hyperref} %<--- Load after everything else
\hypersetup{
  colorlinks=false,
  linkbordercolor=white,
 urlbordercolor=white,
pdfborder={0 0 0}
}
\def\BibTeX{{\rm B\kern-.05em{\sc i\kern-.025em b}\kern-.08em
    T\kern-.1667em\lower.7ex\hbox{E}\kern-.125emX}}
    
\begin{document}
\history{Date of publication xxxx 00, 0000, date of current version xxxx 00, 0000.}
\doi{10.1109/ACCESS.2017.DOI}

%\title{Hand Pose Estimation Based Dynamic Korean Sign Language Recognition Using Attention-based Neural Network}

%\title{EMG Sensor Based Hand Gesture Recognition  using Multi-Stream Deep Learning Based Spatial-Temporal Feature Enhancement  }
%\title{Hierarchical Feature Extraction for Enhanced Spatial-Temporal Dynamics in sEMG-Based Hand Gesture Recognition}
%\title{Hand Gesture Recognition via Multi-Stream Spatial-Temporal EMG Feature Enhancement}
%\title{Hand Gesture Recognition Through sEMG Signals Utilizing Multi-Stream Based Time-Varying Feature Enhancement Approach}

\title{Electromyography-Based Gesture Recognition: Hierarchical Feature Extraction for Enhanced Spatial-Temporal Dynamics}

%\title{Electromyography-Based Gesture Recognition with xAI: Hierarchical Feature Extraction for Enhanced Spatial-Temporal Dynamics}
 \author{\uppercase{Jungpil Shin\orcidicon{0000-0002-7476-2468}\authorrefmark{1}},(Senior IEEE Member), \uppercase{Abu Saleh Musa Miah\ \orcidicon{0000-0002-1238-0464} \authorrefmark{2}},  \uppercase{Sota Konnai \authorrefmark{3}},\uppercase{Shu Hoshitaka \authorrefmark{4}},\uppercase{PANKOO KIM  \authorrefmark{2}}}
 \address[1,2,3]{School of Computer Science and Engineering, The University of Aizu, Aizuwakamatsu, Japan (e-mail:musa@u-aizu.ac.jp)}
  \address[4]{Department of Computer Engineering, Chosun University, Gwangju 61452, South Korea }
  \tfootnote{This research was supported by the "Technology Commercialization Collaboration Platform Construction" project of the Innopolis Foundation (Project Number: 1711177250) and the Competitive Research Fund of The University of Aizu, Japan}
\markboth
{....}
{This paper is currently under review for possible publication in IEEE Access.}
\corresp{Jungpil Shin (jpshin@u-aizu.ac.jp), and Pankoo Kim (pkkim@chosun.ac.kr)}

\begin{abstract}
Hand gesture recognition using multichannel surface electromyography (sEMG) is challenging due to unstable predictions and inefficient time-varying feature enhancement. To overcome the lack of signal-based time-varying feature problems, we propose a lightweight squeeze-excitation deep learning-based multi-stream spatial-temporal dynamics time-varying feature extraction approach to build an effective sEMG-based hand gesture recognition system. 
Each branch of the proposed model was designed to extract hierarchical features, capturing both global and detailed spatial-temporal relationships to ensure feature effectiveness. The first branch, utilizing a Bidirectional-TCN (Bi-TCN), focuses on capturing long-term temporal dependencies by modelling past and future temporal contexts, providing a holistic view of gesture dynamics. The second branch, incorporating a 1D Convolutional layer, separable CNN, and Squeeze-and-Excitation (SE) block, efficiently extracts spatial-temporal features while emphasizing critical feature channels, enhancing feature relevance. The third branch, combining a Temporal Convolutional Network (TCN) and Bidirectional LSTM (BiLSTM), captures bidirectional temporal relationships and time-varying patterns. Outputs from all branches are fused using concatenation to capture subtle variations in the data and then refined with a channel attention module, selectively focusing on the most informative features while improving computational efficiency. The proposed model was tested on the Ninapro DB2, DB4, and DB5 datasets, achieving accuracy rates of 96.41\%, 92.40\%, and 93.34\%, respectively. These results demonstrate the system’s capability to handle complex sEMG dynamics, offering advancements in prosthetic limb control and human-machine interface technologies with significant implications for assistive technologies.\end{abstract}
\titlepgskip=-15pt
\maketitle
%\textcolor{red}{response file}\url{https://docs.google.com/document/d/1ASSlq0EcUaRFdf2qexh4v1OToYCx_agr/edit}
\section{Introduction}\label{sec1}
\label{sec1}
Human activity recognition (HAR) has become a focal point of global research, driven by advancements in human-computer interaction technologies \cite{miah_reivew2024methodological}. Within this field, hand gesture recognition plays a pivotal role, enabling transformative applications such as prosthetic hand control and sign language interpretation. The emergence of wearable sensor technologies, particularly surface electromyography (sEMG) sensors, has revolutionized these systems by offering exceptional flexibility and precision in capturing muscle activity, thereby enhancing the accuracy and reliability of gesture recognition solutions \cite{miah2024effective_EMG_mta}. These sensors noninvasively record sEMG signals, providing valuable data that reflect motor intent and limb movement, making them highly effective for hand gesture recognition. sEMG-based hand gesture recognition is increasingly being adopted in human-machine interaction (HMI) systems. By capturing the electrical activity of muscles, these systems enable precise and intuitive control of devices like prosthetic limbs and rehabilitation tools. Unlike vision-based methods, such as RGB or depth cameras, sEMG systems are unaffected by lighting conditions or visual obstructions, making them ideal for use in low-light environments or when privacy is a concern \cite{montazerin2023transformer}.
A significant advantage of sEMG systems is their independence from visual data, allowing applications in scenarios where vision-based methods fall short. For instance, amputees can use sEMG signals from forearm muscles to control prosthetic devices, enabling natural and lifelike movements \cite{10453986_Zbinden}. Rehabilitation programs also benefit from sEMG systems by tracking motor recovery progress and delivering targeted therapy for patients with physical impairments. Furthermore, in gaming, sEMG systems provide an immersive experience, allowing users to control virtual environments through hand gestures without relying on external cameras \cite{esposito2020piezoresistive}. In addition, sEMG systems excel in environments where vision-based methods struggle, such as underwater robotics, cluttered indoor spaces, or areas with privacy concerns. Their ability to directly measure muscle activity ensures accurate gesture recognition even under challenging conditions, further solidifying their importance in diverse applications. To enhance EMG-based hand gesture recognition, there are many laboratory publicly open EMG datasets where the Ninapro database  \cite{atzori2014electromyography,atzori2015characterization} is a considerable benchmark dataset, which is available under the names of DB1-DB9.  The Ninapro dataset offers kinematic and sEMG signals for 52 finger, hand, and wrist movements, replicating real-world conditions to support the development of advanced DNN-based recognition frameworks.

 \subsection{Current EMG-Based Hand Gesture Recognition Systems and Their Challenges}
 
 Electromyography (EMG)-based hand gesture recognition systems rely on electrodes placed on the arm and forearm to capture surface EMG (sEMG) signals. There are many researchers who have been working to develop ML-based HGR systems using traditional feature extraction and ML approaches. They first processed to extract features and classified using ML techniques such as $k$-nearest neighbours ($k$NN), support vector machines (SVM) \cite{miah2024effective_EMG_mta}, Artificial Neural Network (ANN)\cite{prakash2025optimized}, and random forests (RF) \cite{pizzolato2017comparison_DB4_DB5}. While these methods have shown promise, they face significant challenges in practical applications, particularly in achieving high performance under real-world conditions. One major challenge is signal variability caused by changes in muscle conditions, differences in movement dynamics, user-specific neural control, and inconsistencies in electrode placement \cite{9422807}. Additionally, the time-series nature of sEMG signals and the need to process large datasets hinder the robustness and generalization of traditional ML methods \cite{rahimian2020xceptiontime,rahimian2019semg,chen2020hand,wei2019surface,hu2018novel,ovadia2024classification}. 
To address these issues, researchers have increasingly adopted deep learning (DL) approaches \cite{ovadia2024classification,li2025deep,pinzon2019cnn,geng2016gesture}. Atzori et al. \cite{ovadia2024classification} employed convolutional neural networks (CNNs), achieving 60.27\% $\pm$ 7.7\% accuracy on the Ninapro DB1 and DB2 datasets. Wei et al. \cite{wei2019surface} and Hu et al. \cite{hu2018novel} further improved accuracy to 82.95\% on the DB1 dataset. Du et al. \cite{du} addressed EMG signal variations across sessions using a deep-learning-based domain adaptation framework, validated on multiple datasets, including NinaPro, CSL-HDEMG, and CapgMyo, showcasing its effectiveness in improving classification performance.
Deep neural networks (DNNs) have become increasingly popular for sEMG-based systems, with researchers evaluating their models on the Ninapro DB1–DB9 datasets \cite{miah2023dynamic_Ieee,miah2024hand_multiculture,miah2024spatial_ppa,miah2024sign_largeScale_dataset,rahimian2019semg,rahimian2020xceptiontime,chen2020hand,wei2019surface,ovadia2024classification}. However, these models often require extensive training data, limiting their real-time applicability. To address data limitations, transfer learning has been employed to improve generalization for new users \cite{catallard2019deep}, as demonstrated by Pizzolato et al. using the Ninapro DB5 dataset \cite{pizzolato2017comparison_DB4_DB5}. Temporal convolutional networks (TCNs) have emerged as a promising solution to enhance temporal feature extraction \cite{Tsinganos2019,betthauser2019stable}. Tsinganos et al. \cite{Tsinganos2019} demonstrated a 5\% accuracy improvement using optimized TCN configurations.  
In recent years, advancements in technology and computational resources have enabled researchers to combine TCN with other technologies such as Ma et al. \cite{ffcslt} integrated CNN-SE-LSTM-TCN (FFCSLT) network for traffic police gesture recognition, achieving high accuracy on two datasets by combining diverse deep networks with SE blocks for feature selection and signal transformation. Tsinganos et al. \cite{tsinganos} introduced the TCN model, while Du et al. \cite{du2024} presented the TCN-LSTM model, achieving 89.76\% and 93.78\% accuracy, respectively, on Ninapro DB1, demonstrating the effectiveness of TCN-based approaches for sEMG gesture recognition.  Rahmani et al. \cite{9422807} introduced the FS-HGR model, combining TCN with attention mechanisms, achieving improved accuracy on the Ninapro dataset. Despite these advancements, challenges such as data dependency, user variability, and real-time adaptability persist. Recently, researchers have explored hybrid architectures combining CNN, Bi-LSTM, and TCN to further improve temporal feature recognition \cite{CNN_TCN,LSTM-TCN, shin2024hand_emg}. Addressing these challenges will be pivotal in making EMG-based hand gesture recognition systems more reliable and practical for applications such as prosthetic control and rehabilitation.

\subsection{Motivition}
Accurate hand gesture classification using multichannel surface electromyography (sEMG) signals holds significant potential across various domains. A robust gesture recognition model is essential for practical applications, as sEMG signals can vary depending on factors such as context and device. However, existing architectures often face challenges in maintaining high performance across diverse conditions, primarily due to unstable predictions and difficulties in capturing time-varying features effectively. The spatial-temporal dynamics of sEMG signals are critical for applications such as prosthetic control, rehabilitation, and human-machine interaction. Yet, variability caused by muscle condition changes, inconsistent electrode placement, and dynamic movement patterns limits the robustness, adaptability, and generalization of current systems. Many existing methods fail to adequately enhance time-varying features, and while neural network combinations like CNN-TCN and LSTM-TCN have shown promise, their potential for sEMG-based hand gesture recognition remains underexplored. To address these gaps, we propose a lightweight, scalable multi-stream deep learning architecture that extracts hierarchical spatial-temporal features using advanced techniques like Bidirectional-TCN (Bi-TCN), separable CNN, and Squeeze-and-Excitation (SE) blocks. This study focuses on classifying 52 hand gestures from the Ninapro dataset, a publicly available benchmark, aiming to achieve accuracy surpassing previous studies. By experimenting with multiple datasets, we mitigate performance discrepancies caused by variations in sEMG equipment used during data collection. Our goal is to establish a high-performance framework that advances state-of-the-art sEMG-based hand gesture recognition. By prioritizing the time-varying nature of sEMG signals, this approach enhances recognition accuracy, robustness, and computational efficiency, paving the way for practical and reliable sEMG-based hand gesture recognition systems.

\subsection{The Goal and Scope of the Study}
The contributions of the proposed study are given below: 
\begin{itemize} 
\item \textbf{Three-Stream Architecture for Temporal and Spatial Feature Extraction:} We designed a three-stream network architecture, where each stream is specialized in extracting distinct aspects of the sEMG signal. The first stream applies bidirectional-TCN to capture both forward and backward temporal relationships, strengthening the model’s ability to process complex temporal patterns.
The second stream combines 1D convolution, separable CNN, and Squeeze-and-Excitation (SE) Block to extract spatial and temporal features while improving the efficiency and quality of feature extraction. The third stream utilizes Temporal Convolutional Networks (TCN) followed by BiLSTM to capture temporal dependencies and extract compelling temporal features.

\item \textbf{Feature Fusion and Dimensionality Reduction for Improved Classification:} The features from the three streams are concatenated to form a comprehensive, multi-faceted feature representation. A channel attention module is then employed to selectively reduce the dimensionality of these features, improving computational efficiency and ensuring the most relevant features are used for classification. This fusion strategy ensures the model leverages complementary temporal and spatial information for enhanced gesture recognition accuracy.
\item \textbf{Novelty in Temporal-Spatial Feature Integration:} The model’s novelty lies in the integration of temporal and spatial feature extraction through specialized modules in each stream. By using TCN in the first stream, we capitalize on the long-term dependencies within the sEMG signal. The hybrid LSTM-TCN module in the second stream helps capture more intricate temporal relationships, while the Bidirectional-TCN in the third stream ensures comprehensive learning of temporal patterns. This fusion allows for an enhanced understanding of the dynamic and static aspects of hand gestures.
\item \textbf{Model Validation and Robustness Evaluation:} We validated the proposed model by conducting experiments on the Ninapro DB2, DB4, and DB5 datasets, achieving accuracies of 96.41\%, 92.40\%, and 93.34\%, respectively. These results demonstrate the model’s robustness and effectiveness across different datasets. The performance of the proposed model is significantly higher compared to existing methods, showcasing its ability to handle diverse hand gesture recognition tasks in the context of sEMG-based systems.
\end{itemize}

The structure of the paper is outlined as follows: Section \ref{sec2} offers an overview of related approaches to gesture recognition.
In Section  \ref{meth:dataset}, we provide a detailed description of the dataset employed in this study.
The TCN-based model proposed in this paper is introduced in Section  \ref{sec4}, followed by the presentation of results % and a discussion
in Section  \ref{res:result}.
Lastly, Section \ref{sec6} summarizes the conclusions and outlines potential avenues for future work.

\section{Related Work} \label{sec2}
In recent years, gesture recognition using sEMG data has attracted attention in many fields, including medicine, exercise science, engineering, and prosthetic limb control\cite{miah_reivew2024methodological,10453986_Zbinden}. %basic and RGB based work
Wearable sensor-based methods offer distinct advantages, including faster recognition, higher accuracy, and robustness across various environments. The growing interest in this field is fueled by advancements in machine learning (ML), deep learning (DL), and rehabilitation technologies. Traditional ML techniques such as Linear Discriminant Analysis (LDA) and Support Vector Machines (SVM) have been applied to classify hand gestures derived from sEMG signals. While myoelectric control using conventional pattern recognition techniques has been extensively studied in academic research, these methods are rarely adopted in commercial applications due to a gap between real-world requirements and existing solutions \cite{castellini2014proceedings}. Hand gesture recognition primarily evolves from two key data modalities: the data captured from sensor signals and the associated motion features. Liu et al. \cite{liu2021neuropose} leveraged  3D finger poses, which is non-trivial since signals from multiple fingers superimpose at the sensor in complex patterns. Still, the accuracy was constrained by the limited range of features these sensors could capture. In another study, Prakash et al. \cite{prakash2025optimized} integrated data from multiple sensors and achieved an impressive 85.90\% recognition accuracy for the GrabMyo dataset using SVM, similar to the RF methods. These advancements demonstrate the potential of wearable sensor-based systems while highlighting the need for further improvements in real-world applications.
%deep learning based system, RNN, LSTM, 3DCNN
Deep learning techniques have emerged as powerful tools for sEMG-based hand gesture recognition, addressing many of the limitations of traditional machine learning methods. Convolutional neural networks (CNNs) have been widely used among the foundational approaches. Li et al. \cite{li2025deep} developed an end-to-end deep learning framework using CNNs to classify hand gestures, achieving higher accuracy compared to traditional ML techniques. Pinzón-Arenas et al. \cite{pinzon2019cnn} leveraged power density maps of sEMG signals as CNN inputs, effectively recognizing six gestures. Similarly, Geng et al. \cite{geng2016gesture} used high-density sEMG images as CNN inputs, achieving an 89.30\% recognition rate for eight gestures. Despite their effectiveness, CNN-based methods often require extensive training data and struggle with temporal dependencies in sEMG signals \cite{miah2023dynamic_Ieee, miah2024hand_multiculture,miah2024spatial_ppa,miah2024sign_largeScale_dataset}.
While existing Deep Neural Network (DNN) techniques show impressive performance on unseen repetitions, they struggle with repetitions that haven't been thoroughly examined \cite{srinivasan2021optimization}. For example, \cite{catallard2019deep} evaluates the average accuracy for 10 subjects from the Ninapro DB5 dataset, considering various numbers of training repetitions (each repetition represents 5 seconds of data). It is observed that accuracy decreases, and the model faces difficulties when the repetitions are not well explored. Specifically, the reported accuracies for one to four training repetitions are 49.41\% ± 5.82\%, 60.12\% ± 4.79\%, 65.16\% ± 4.46\%, and 68.98\% ± 4.46\%, respectively. Chattopadhyay et al. proposed a domain adaptation method that aligns the original and target data into a common domain while maintaining the probability distribution of the input data \cite{chattopadhyay2011topology}. To overcome these challenges, transfer learning (TL) has been employed to expedite training and improve generalization for new subjects. TL-based algorithms, as presented by Catallard et al. \cite{catallard2019deep}, utilized pre-trained CNN models to transfer knowledge across diverse subjects, achieving commendable accuracy on the Ninapro DB5 dataset. Pizzolato et al. \cite{pizzolato2017comparison_DB4_DB5} also applied TL, demonstrating its effectiveness in improving recognition accuracy for sEMG signals collected using the Myo armband. This database includes data from 10 subjects with intact limbs, and a high accuracy was achieved. Advanced models like 3D CNNs have been introduced to address spatio-temporal feature extraction. Jiang et al. \cite{jiang2021efficient} developed a 3D CNN module to simultaneously capture spatial and temporal information. Hybrid architectures have also gained traction. Researchers have been working to improve it within sequential feature-based models, such as recurrent neural networks (RNNs) and long short-term memory (LSTM) networks, which have shown promise in capturing temporal dependencies. Ketykó et al. \cite{ketyko2019domain} proposed a two-stage domain adaptive framework using RNNs, achieving a 90.70\% recognition rate for 12 gestures. Quivira et al. \cite{quivira2018translating} employed LSTMs for continuous hand pose recognition, while Samadani \cite{samadani2018gated} demonstrated enhanced performance using bidirectional LSTMs. Moreover,  Chen et al. \cite{chen2020capsnet} proposed CapsNet to extract more spatial-temporal enhanced features, aiming to extract multi-angle range discrete features. Chen et al. \cite{chen2021deep} implemented sub-networks for processing sensor data and combined their outputs with an LSTM network to extract sequential information. Recent research emphasizes that deep learning can significantly enhance generalization, robustness, and accuracy over traditional machine learning methods, which is essential for practical applications. To alleviate the effort involved in manually designing features, scientists have started using deep learning models to recognize sEMG signals. For instance, Li et al. \cite{li2025deep} developed an end-to-end deep learning framework with convolutional neural networks (CNNs) to classify hand movements from sEMG signals, demonstrating higher accuracy compared to conventional machine learning techniques. 
Tuncer and Alkan \cite{tuncer2022cnn} utilized hybrid deep learning frameworks, and Bai et al. \cite{prabhavathy2024surface_CNN_LSTM} integrated CNN and LSTM models to achieve promising results for multi-channel sEMG signal recognition. Temporal Convolutional Networks (TCNs) have emerged as a robust alternative for handling temporal features in sEMG-based systems. Tsinganos et al. \cite{Tsinganos2019} demonstrated a 5\% improvement in accuracy using optimized TCN configurations and enhancement of the TCN used by \cite{betthauser2019stable}.  However, existing systems with TCN modules still face challenges in achieving high-performance accuracy. Recently, some researchers in various domains have integrated CNN with LSTM and TCN to enhance the recognition of temporal features \cite{CNN_TCN,LSTM-TCN}.  We have not found any research combining these features to recognize sEMG-based hand gesture signals. Rahmani et al.~proposed the FS-HGR model to improve the performance accuracy of the Ninapro sEMG dataset.~They mainly employed a combination of the TCN and attention model, and they integrated 3 TCN and four attention blocks \cite{9422807}. Joseph et al. \cite{betthauser2019stable} utilized TCN modules to refine muscle activity patterns. Rahmani et al. \cite{9422807} combined TCN with attention mechanisms, introducing the FS-HGR model, which integrated three TCN and four attention blocks to achieve higher accuracy on the Ninapro dataset. Researchers have recently explored hybrid architectures combining CNN, Bi-LSTM, and TCN to improve temporal feature recognition \cite{CNN_TCN, LSTM-TCN}. To overcome these challenges, we propose an effective time-varying feature enhancement module to classify sEMG-based hand gestures. 
The proposed method introduces a robust three-stream architecture designed for effective temporal and spatial feature extraction from sEMG signals. The first stream employs Bidirectional-TCN to analyze both forward and backward temporal relationships, while the second stream integrates 1D convolution, separable CNN, and Squeeze-and-Excitation (SE) blocks to enhance spatial and temporal feature representation. The third stream utilizes TCN followed by BiLSTM to effectively capture temporal dependencies. To improve classification accuracy, features from the three streams are fused and processed through a channel attention module, ensuring the most relevant features are selected while reducing computational complexity. Validation on the Ninapro DB2, DB4, and DB5 datasets demonstrates high accuracies (96.41\%, 92.40\%, and 93.34\%, respectively), highlighting the model's robustness and effectiveness.
The method's novelty lies in its integration of advanced temporal-spatial feature extraction techniques, leveraging specialized modules for long-term dependencies, intricate temporal relationships, and comprehensive pattern learning, significantly enhancing gesture recognition accuracy and generalizability \cite{atzoribenchmark_DB1}.

\begin{comment}
 Fig.~\ref{fig:cybergloves} shows an example of these glove-integrated sensors.
{\begin{figure}[ht]
\centering
\includegraphics[scale=0.15]{./Images/dataset_cybergloves.jpg}
\caption{Cybergloves to record the dataset}
\label{fig:cybergloves}
\end{figure} }
\end{comment}
\begin{comment}
{\begin{figure*}[ht]
\centering
\includegraphics[scale=0.20]{./Images/dataset_DB1.jpg}
\caption{Sample images of the DB1 dataset}
\label{fig:DB1_dataset}
\end{figure*} }
\end{comment}

\begin{table}[H]
\centering
\caption{Information about the hand gesture sEMG-based datasets DB2, DB4, and DB5} \label{tab:dataset}
\begin{small} % Adjusts the font size for better readability
\begin{tabular}{|l|c|c|c|c|}
\hline
\textbf{Attribute Name} & \textbf{DB2} & \textbf{DB4} & \textbf{DB5} \\ \hline
             
Number of channels                 & 12           & 12           & 16                    \\ \hline
Number of repetitions              & 10           & 6            & 6    \\ \hline
Number of gestures                 & 50           & 52           & 52  \\ \hline
 Number of subjects                 & 40           & 10           & 10                      \\ \hline 
     Number of Male                 & -           & 6           & 8      \\ \hline 
          Number of Female                 & -          & 4           & 2  \\ \hline 

\end{tabular}
\end{small}
\end{table}
{\begin{figure}[ht]
\centering
\includegraphics[scale=0.30]{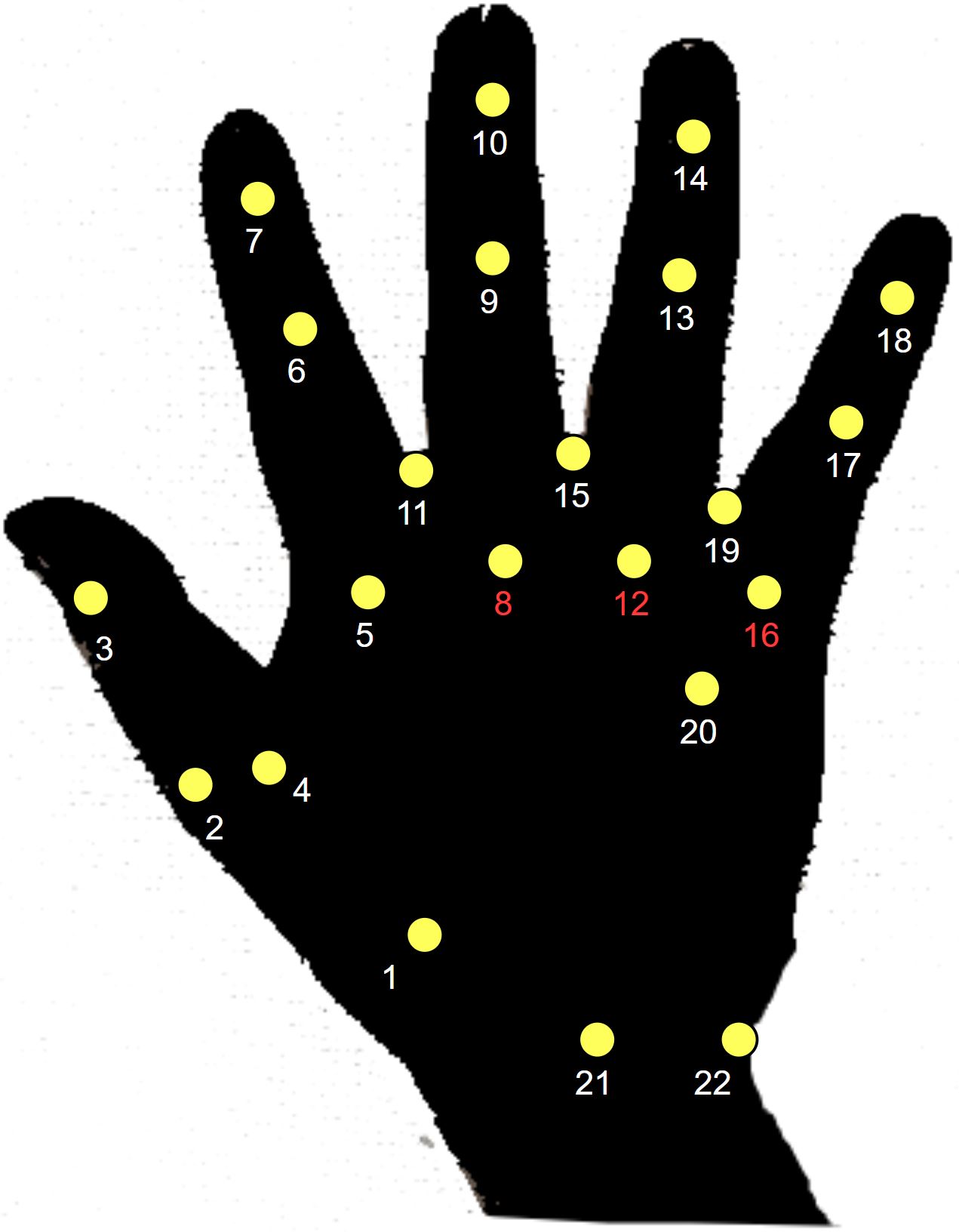}
\caption{Data collection sensor location for the Ninapro DB \cite{atzori2014electromyography,atzori2015characterization,atzori2016deep}}
\label{fig:sensor_placed_onhand}
\end{figure} }

\begin{figure}[ht]
\begin{adjustwidth}{-0cm}{0cm}
\centering
\includegraphics[scale=0.35]{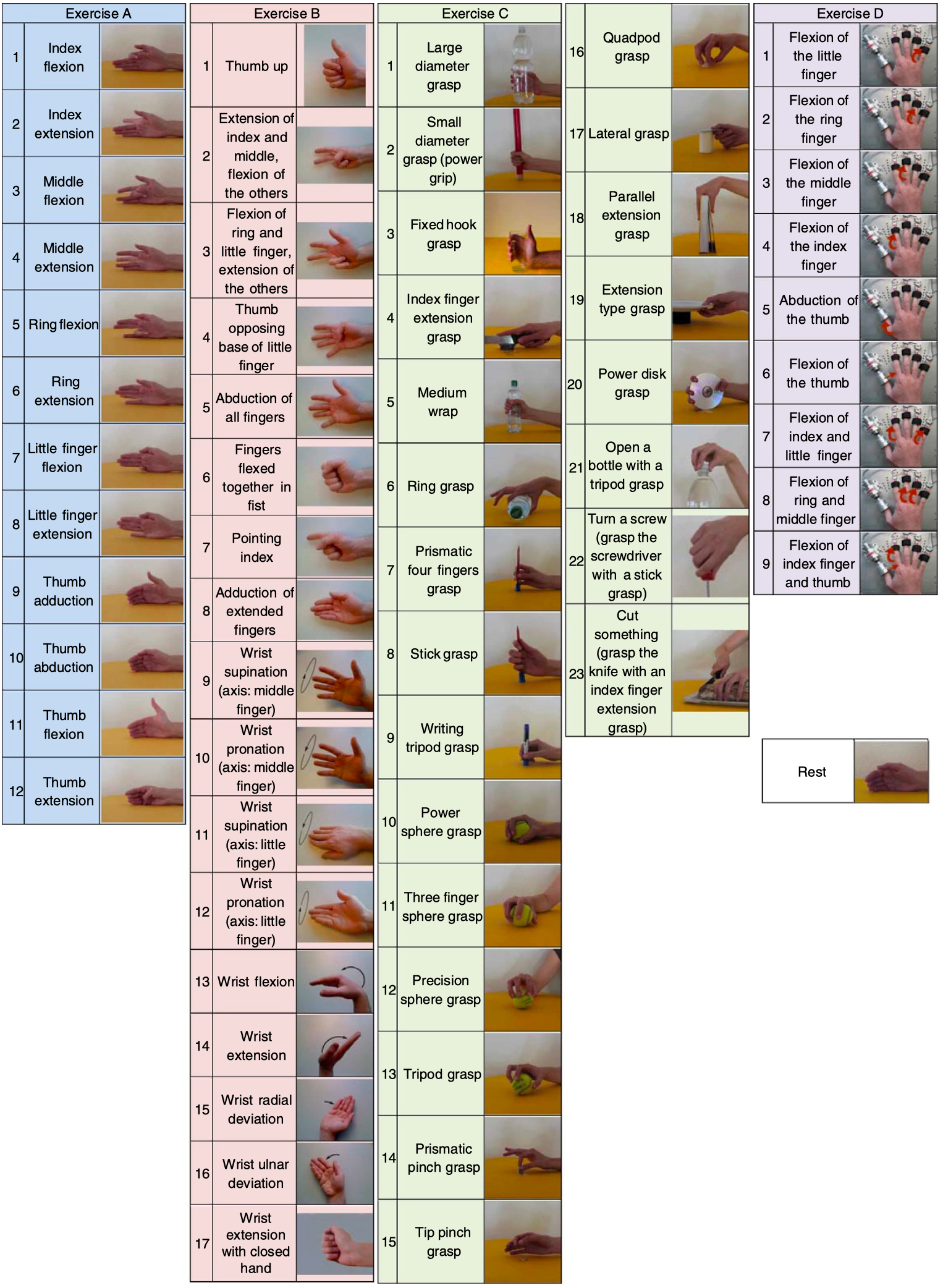}
\caption{Sample images of the hand gesture Ninapro DB2, DB4 and DB5 datasets \cite{atzori2014electromyography,atzori2015characterization,atzori2016deep}}
\label{fig:DB1_dataset}
\end{adjustwidth}
\end{figure} 

% \begin{figure}[ht]
% \centering
% \includegraphics[scale=0.40]{Images/Gestures_1.png}
% \caption{Sample images of the DB4 and DB5 dataset \cite{atzori2014}}
% \label{fig:DB1_dataset}
% \end{figure} 

\section{Dataset Description}
\label{meth:dataset}
In the study, we used three benchmark datasets, including Ninapro DB2, DB4, and DB5, to evaluate the proposed model. We describe each dataset below. Figure \ref{dataset_cybergloves.JPG} shows the sensor location in the hand, and Figure \ref{fig:DB1_dataset} demonstrates the sample image of the hand gesture, and Table \ref{tab:dataset} shows the description of each dataset. 

\subsection{DB2 Dataset}
The Ninapro DB2 dataset includes sEMG, inertial, kinematic, and force data collected from 40 healthy subjects. These subjects performed 49 different hand movements plus a rest position. Detailed information about this dataset is available in the paper by Atzori et al., "Electromyography data for non-invasive naturally-controlled robotic hand prostheses," published in Scientific Data in 2014 \cite{atzori2014electromyography}. The data collection protocol has three main exercises: basic finger movements, grasping and functional tasks, and force patterns. Subjects repeated movements shown on a laptop screen, with each movement lasting 5 seconds followed by a 3-second rest.
Kinematic data were captured using a Cyberglove 2, while force data were recorded with a Finger Force Linear Sensor. The sEMG signals were recorded using 12 Delsys Trigno electrodes placed on the forearm, flexor, and extensor muscles, as well as the biceps and triceps. These signals were sampled at 2 kHz. The dataset also includes force measurements, accelerometer data, and joint angle data from the Cyberglove.

%\subsection*{DB3 Dataset}
\subsection{DB4 Dataset}
The Ninapro DB4 dataset contains sEMG and kinematic data recorded from 10 intact subjects performing 52 hand movements, including the rest position. The dataset is described in detail in the publication by Pizzolato et al., "Comparison of six electromyography acquisition setups on hand movement classification tasks," PLOS One, 2017. The subjects repeated the movements shown in videos on a laptop screen using their right hand. Each movement lasted approximately 5 seconds, followed by 3 seconds of rest. The experiment was divided into three exercises: basic finger movements, isometric and isotonic hand configurations with wrist movements, and functional grasping motions. The sEMG data were captured with 12 Cometa electrodes at a 2 kHz sampling rate. The dataset includes synchronized MATLAB files with variables such as subject details, sEMG signals, cyberglove data (measuring joint angles), movement stimuli, and demographic information like age, gender, weight, and handedness \cite{pizzolato2017comparison_DB4_DB5}.
\subsection{DB5 Dataset}
The Ninapro DB5 dataset, part of the Ninapro series, includes data collected from 10 intact subjects using two Thalmic Myo armbands. Designed to assess electromyography (sEMG) systems, this dataset focuses on hand movement classification tasks and provides a comprehensive examination of various hand and wrist movements. The acquisition protocol involves three main exercises: basic finger movements, isometric and isotonic hand configurations with wrist motions, and functional grasping movements. Each subject repeated 52 distinct movements, performing six repetitions of each, with 5-second movement intervals followed by 3 seconds of rest.
The Myo armbands, featuring 16 active single-differential wireless electrodes, were placed strategically along the arm—one near the elbow and the other closer to the hand, tilted at a 22.5-degree angle. This configuration ensures a broad and uniform muscle mapping while remaining cost-effective. The sEMG signals were sampled at 200 Hz, making this dataset ideal for testing the Myo armbands both individually and together \cite{pizzolato2017comparison_DB4_DB5}.

{\begin{figure*}[ht]
\centering
\includegraphics[scale=0.35]{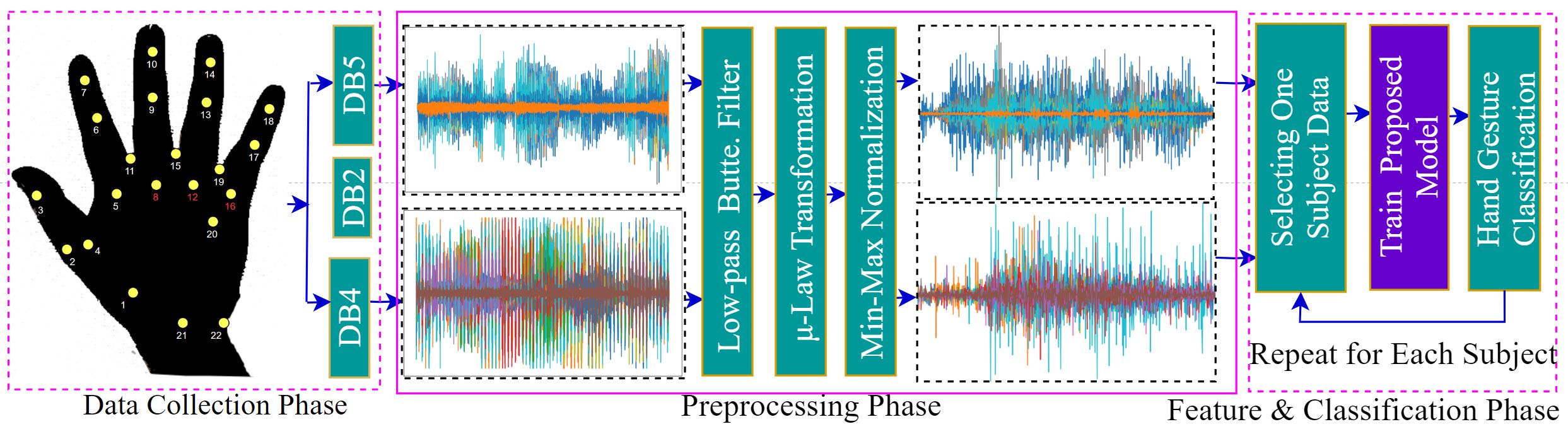}
\caption{The workflow for hand gesture classification using
EMG signal}
\label{fig:preprocessing}
\end{figure*} }
\section{Proposed Method} \label{sec4}
In this study, we propose a novel deep learning-based hand gesture recognition system leveraging surface electromyography (sEMG) signals to address key challenges in gesture recognition, such as unstable predictions and inefficient feature extraction. Figure \ref{fig:preprocessing} demonstrates the workflow architecture of the proposed model, including three phases, namely: data collection phase, preprocessing phase and feature extraction and classification phase. Figure \ref{fig:main_figure} demonstrates the proposed three deep learning models. Our system features a robust three-stream architecture designed to capture hierarchical features, emphasizing both spatial and temporal relationships while incorporating advanced feature fusion and dimensionality reduction techniques. This innovative design enhances classification accuracy and computational efficiency, making it suitable for real-time applications. Inspired by the multistage design of \cite{9422807}, which effectively combined Temporal Convolutional Networks (TCN) with attention mechanisms, our proposed model introduces three specialized streams to comprehensively extract complementary features from sEMG signals:
\begin{itemize}
    \item 
\textbf{Stream 1: Short-and Long-Term temporal relationships:}
In the first stream, we employed a Bidirectional Temporal Convolutional Network (Bi-TCN) to capture long-term temporal dependencies by modelling both past and future contexts. This bidirectional approach enables the model to comprehensively understand gesture dynamics, ensuring a robust representation of sequential patterns in sEMG signals.
\item 
\textbf{Stream 2: Calibrating channel-wise feature with SE} The second stream integrates a 1D Convolutional layer, a separable CNN, and a Squeeze-and-Excitation (SE) block. The 1D Convolutional layer captures local patterns, while the separable CNN extracts complex spatial-temporal dependencies efficiently. The SE block recalibrates feature maps by emphasizing the most informative channels, reducing overfitting and improving feature relevance.
The integration of both CNN and Sep-TCN in this stream allows the model to gain a comprehensive understanding of the complex spatial-temporal relationships in the sEMG data. This branch’s novelty lies in the combination of separable CNN and SE blocks, which offer efficient feature extraction while focusing on critical spatial-temporal interactions in the EMG data.
\item 
\textbf{Stream 3: Intricate time-varying feature extraction}
The third stream leverages the power of TCN to capture long-range temporal dependencies inherent in sequential sEMG data. TCNs are well-suited for learning from sequential patterns, allowing the model to identify intricate temporal features critical for accurate gesture recognition. To further enhance temporal feature extraction, we integrate a Bidirectional LSTM (BiLSTM) layer, which enables the model to capture both forward and backward temporal relationships in the sEMG signal. This combined approach improves the model's understanding of complex temporal dynamics over time.

\item 
\textbf{Feature Fusion and Dimensionality Reduction}
The features extracted from the three streams are concatenated into a unified feature representation. We incorporate a channel attention module to ensure that the most relevant features are selected and to reduce the dimensionality of the combined feature set. This attention mechanism helps focus on the most discriminative features, reducing computational complexity while retaining essential information. The dimensionality-reduced features are then fed into the classification module for gesture recognition.
\item 
\textbf{Classification and Model Evaluation}
The final feature representation, after attention-based dimensionality reduction, is passed through a fully connected (FC) layer for classification. The output of the FC layer is a probabilistic map indicating the predicted gesture. We validate the effectiveness of the proposed system through extensive experiments on the Ninapro DB2, DB4, and DB5 datasets. %The results demonstrate that the model achieves impressive classification accuracy of 95.00\%, 95.00\%, 91.92\%, and 93.05\%, respectively, outperforming existing methods.
The proposed methodology integrates temporal and spatial feature extraction, attention-based feature selection, and multi-stream fusion, providing a comprehensive solution for EMG-based hand gesture recognition. In the following sections, we will describe %the dataset (\ref{meth:dataset}),%
preprocessing (\ref{meth:prep}), multi-stream feature extraction module (\ref{meth:mult1}),  (\ref{meth:mult2}),  (\ref{meth:mult3}), feature concatenation (\ref{meth:feat}), and the classification process (\ref{meth:classification}). The algorithm for the proposed model is outlined in the next section.
\end{itemize}

{\begin{figure*}[ht]
\centering
\includegraphics[scale=0.30]{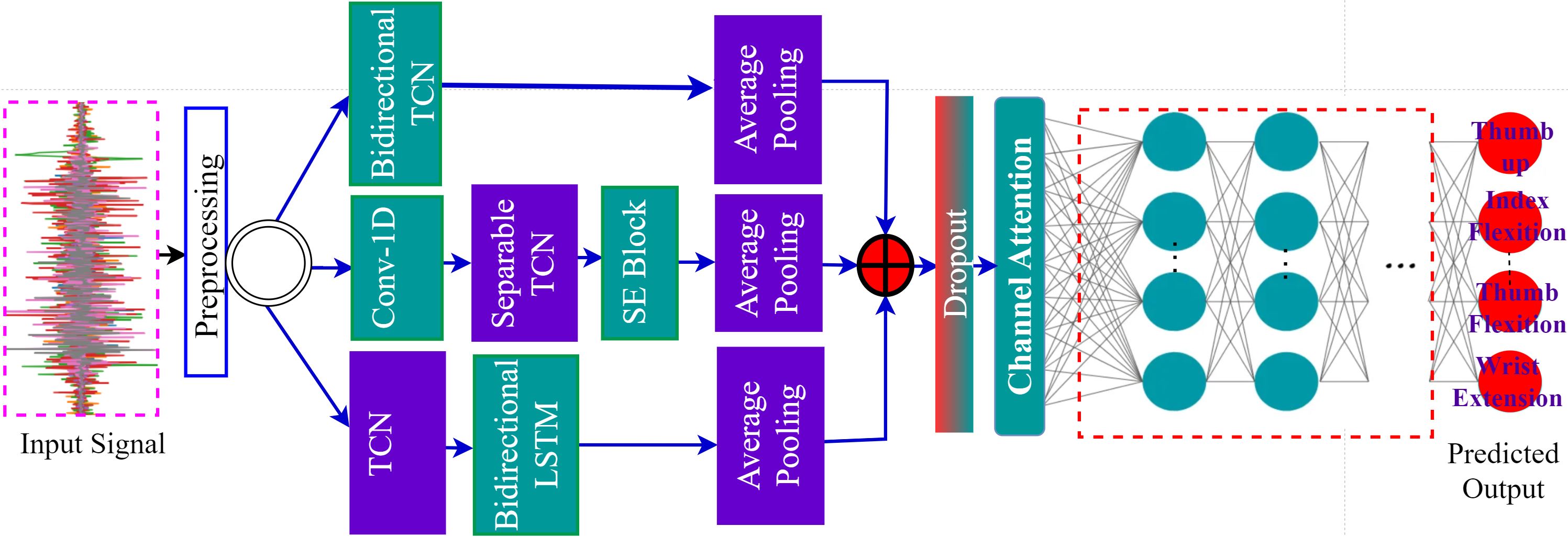}
\caption{The structure of our proposed model}
\label{fig:main_figure}
\end{figure*} }

\subsection{Preprocessing}
\label{meth:prep}
Figure \ref{fig:preprocessing} demonstrates the workflow architecture of the proposed model, including preprocessing. In the study, we used Ninapro DB2, DB4 and DB5 datasets, which are widely used for EMG-based hand gesture recognition or myoelectric control. These datasets feature almost 50 gestures from 10-40 subjects, each repeated 6-10 times and recorded at 200-2,000 Hz. %This study emphasizes 23 complex grasping and functional movements, which is ideal for validating the model's effectiveness.

\subsubsection{Standardization}
Electromyography (EMG) signals exhibit significant variability due to differences in sensor placement, muscle activity, and individual subject characteristics, making standardization essential for ensuring consistency and comparability across features. Z-score standardization addresses this need by centering the data around a mean of 0 and scaling it to have a standard deviation of 1. This process involves subtracting the mean (\(\mu\)) of the original data from each data point (\(X\)) and dividing the result by the standard deviation (\(\sigma\)), transforming the data into a normalized form that eliminates magnitude differences.

For EMG datasets, standardization is particularly beneficial as it minimizes the influence of range disparities and ensures that each feature contributes equally during model training. This helps the learning process focus on identifying meaningful patterns rather than being dominated by scale differences. By removing biases caused by varying magnitudes, Z-score standardization improves the interpretability and generalizability of models working with dynamic and complex EMG signals.

Additionally, Z-score standardization enhances the efficiency of optimization algorithms by facilitating faster convergence during training. It suppresses irrelevant variations and amplifies the informative features, making it easier for the model to capture essential patterns in the data. This preprocessing step not only reduces training time but also ensures robust performance, making it an indispensable technique for EMG-based hand gesture recognition tasks.

\subsubsection{Data Augmentation with Gaussian Noise}
To address the issue of small sample sizes for certain gestures (which could lead to overfitting), we apply Gaussian noise to expand the dataset. Gaussian noise with a mean of 0 and a variance of 0.1 is added to the data. This augmentation method preserves the frequency, spatial, and energy components of the original signals while altering the signal amplitude.

The Gaussian noise is mathematically defined by the probability density function \(P(z)\) of a random variable \(z\), which follows the normal distribution:
\[
P(z) = \frac{1}{\sqrt{2\pi \sigma}} e^{-\frac{(z - \mu)^2}{2\sigma^2}}
\]
where,  \(\mu\) is the mean, \(\sigma\) is the standard deviation of the noise. The noise is added in such a way that it does not alter the overall characteristics of the signal, ensuring the time-frequency domain remains consistent.
\subsubsection{Data Slicing}
The dataset is sliced into a $10 × 500$ format, where each slice consists of 500 data points representing a single time window. After preprocessing, a total of 6,102 samples are obtained for training. Each gesture's data is extracted from the continuous signal and organized into manageable chunks (time-series segments) to be used as input for model training.
\subsubsection{Train-Validation-Test Split}
The dataset is then divided into training, validation, and test sets using a 6:2:2 ratio to ensure the model is evaluated on unseen data and to prevent overfitting.
{\begin{figure}[ht]
\centering
\includegraphics[scale=0.20]{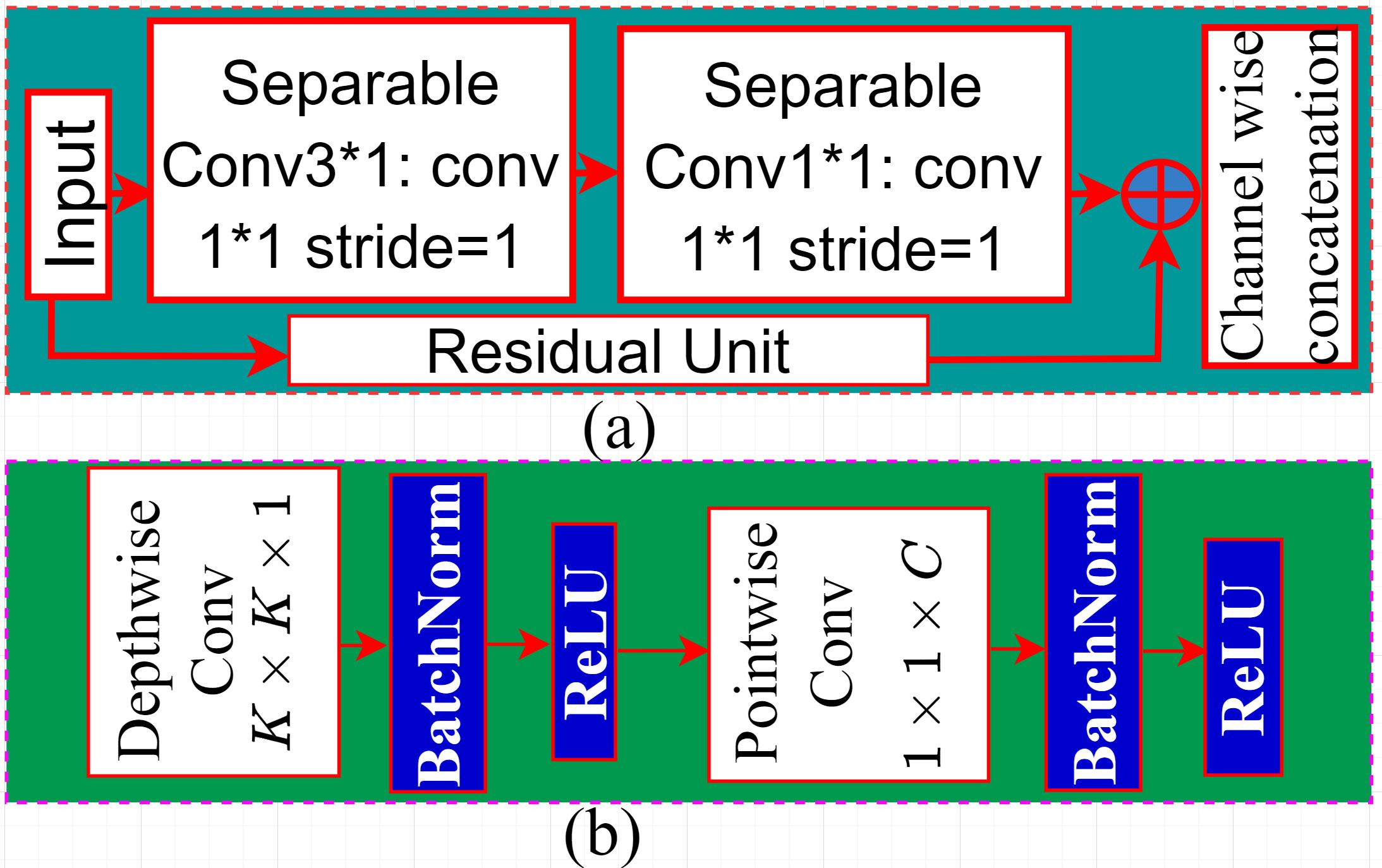}
\caption{Architecture of the Separable CNN\cite{ma2024ffcslt_main_traffic_emg}.}
\label{fig:sep_tcn}
\end{figure} }
\subsection{Stream-1: Bidirectional TCN}
\label{meth:mult1}
We designed a three-stream network architecture, where each stream is specialized in extracting distinct aspects of the sEMG signal. The first stream applies Bidirectional-TCN to capture both forward and backward temporal relationships, strengthening the model’s ability to process complex temporal patterns.

\subsubsection{TCN}
Although Recurrent Neural Networks (RNNs) are effective at establishing sequence-to-sequence correlations, they process one step at a time, which can lead to inefficiencies, particularly when handling long sequences. This limitation results in significant computational overhead. Van den Oord et al. \cite{oord2016wavenet_36} introduced causal convolution as a new approach for modelling time sequences, achieving promising results in speech recognition tasks. Building on this idea, Bai et al. \cite{bai2018empirical_37} proposed the Temporal Convolutional Network (TCN), which leverages the power of convolutional structures. Their results showed that simple convolutional architectures outperformed traditional RNNs in terms of both performance and memory length across a variety of tasks and datasets. The key advantage of TCNs is their ability to capture long-range temporal dependencies more effectively than RNNs.
TCNs employ dilation causal convolution \cite{yuan2021quality_38} to capture temporal dependencies across sequences. By setting the dilation factor, the dilation convolution skips a specified number of steps, thus increasing the receptive field. This technique enables the network to capture long-range dependencies in the sequence, as illustrated in Fig. \ref{fig:tcn}. The TCN approach is highly parallelizable, making it more computationally efficient than RNN-based methods, particularly for tasks with large sequences like hand gesture recognition from sEMG signals.
For a given input signal, the TCN captures temporal dependencies through dilated convolutions. 
Let \( x(t) \in \mathbb{R}^{d} \) represent the input sequence at time step \( t \), where \( d \) is the input dimensionality (e.g., the number of EMG channels). The signal is passed through multiple layers of convolutions with dilation.
\paragraph{Causal Convolution}
The core idea of causal convolution in TCN is that the output at time \( t \), \( y(t) \), is only influenced by the current and past inputs, but not future ones.
\begin{equation}
y(t) = \sum_{k=0}^{K-1} w_k x(t-k)
\label{eq:causal_convolution}
\end{equation}
Where, \( y(t) \) is the output at time \( t \), \( w_k \) is the kernel filter at position \( k \), \( K \) is the kernel size.
\paragraph{Dilated Convolution}
In dilated convolution, the kernel skips \( d-1 \) positions for each dilation level \( d \), which allows the receptive field to increase exponentially without increasing the number of parameters. For a given dilation rate \( d \), the operation is modified as:
\begin{equation}
y(t) = \sum_{k=0}^{K-1} w_k x(t - d \cdot k)
\label{eq:dilated_convolution}
\end{equation}
Where \( d \) is the dilation factor, the dilation increases the receptive field by skipping steps.
\paragraph{Stacked Residual Blocks in TCN}
A TCN is built from stacked residual blocks. The input \( x(t) \) to each block undergoes the convolution operation described above, and then a residual connection is added:
\begin{equation}
z(t) = x(t) + F(x(t))
\label{eq:residual_block}
\end{equation}
Where \( F(x(t)) \) is the result of the dilated convolutional block. This helps avoid vanishing gradients and improves training stability. The output of the residual block is:

\begin{equation}
y(t) = \text{ReLU}(z(t))
\label{eq:relu_output}
\end{equation}

Finally, the output \( y(t) \) is passed to subsequent layers.
\subsubsection{ Bidirectional Temporal Convolutional Network (Bi-TCN)}
In the Bidirectional TCN (Bi-TCN), the sequence is processed in both forward and backward directions. This allows the model to learn dependencies from both the past and future at each time step.

Let \( x(t) \) be the input signal at time step \( t \). For a Bidirectional TCN, we have two streams for forward \ref{eq:forward_tcn} and backwards \ref{eq:backward_tcn}.
\begin{equation}
y_f(t) = \sum_{k=0}^{K-1} w_k x(t-k)
\label{eq:forward_tcn}
\end{equation}
\begin{equation}
y_b(t) = \sum_{k=0}^{K-1} w_k x(t+k)
\label{eq:backward_tcn}
\end{equation}
The outputs from both the forward and backward directions are combined, typically through concatenation or addition:
\begin{equation}
y(t) = \text{concat}(y_f(t), y_b(t)) \quad \text{or} \quad y(t) = y_f(t) + y_b(t)
\label{eq:bi_tcn_output}
\end{equation}
Where \( y(t) \) is the final output at time \( t \) after combining both directions.
Given that the EMG signal is a time-series input, the feature extraction process can be summarized as follows:
Let the raw EMG signal be represented as a 3D tensor \( X \in \mathbb{R}^{N \times T \times C} \), where \( N \) is the batch size,\( T \) is the number of time steps (signal length), \( C \) is the number of channels (sEMG electrodes).
The signal is passed through multiple TCN layers (with causal and dilated convolutions) to capture the temporal dependencies. For each layer, the input \( X \) is transformed using the formula for causal dilated convolutions in equation \(\ref{eq:dilated_convolution}\).
In Bi-TCN, the sequence is processed in both forward and backward directions to capture information from both past and future, increasing the contextual understanding of each time step as described in equations \(\ref{eq:forward_tcn}\) and \(\ref{eq:backward_tcn}\).
The output from the forward and backward directions are combined through concatenation or addition as shown in equation \(\ref{eq:bi_tcn_output}\). 
For a signal \( X(t) \in \mathbb{R}^{C} \) (at time \( t \)), the feature extraction process can be represented of the feature of TCN output\ref{eq:tcn_output}, Bi-TCN output \ref{eq:bi_tcn_output}.  
    \begin{equation}
    F_{TCN}=y(t) = \text{TCN}(X(t)) = \sum_{k=0}^{K-1} w_k X(t - d \cdot k)  \quad 
    \label{eq:tcn_output}
    \end{equation}

    \begin{equation}
    y_b(t) = \text{TCN}(X(t))
    \label{eq:tcn_output_2}
    \end{equation}
Where y(t) is the output of the TCN, $y_f(t)$ represents forward features, and $y_b(t)$ represents backward features. Finally, combine the forward and backward features and then produce the final feature of Bi-TCN as shown in Equation \ref{eq:bi_tcn_output}.  
    \begin{equation}
    F_{Stream-1}=y(t) = \text{concat}(y_f(t), y_b(t)) \quad 
    \label{eq:bi_tcn_output}
    \end{equation}
     
This equation-based approach highlights how the TCN and Bi-TCN models handle temporal dependencies in EMG signal data, using dilated convolutions to capture long-range dependencies and bidirectional processing to capture both past and future information. The extracted features are then fused and classified to recognize hand gestures in real-time applications. The output of the BiTCN is fed into the average pooling layer and produces the feature for $Feature_{branch-1}$

\subsection{Stream-2: CNN-1D, Separable-CNN and SE Block}
\label{meth:mult2}
The second stream combines 1D convolution, separable CNN, and Squeeze-and-Excitation (SE) Block to extract spatial and temporal features while improving the efficiency and quality of feature extraction.

\subsubsection{Depthwise separable convolutional network (DSCN)}
Depthwise separable convolutional networks (DSCNs) are highly effective in extracting meaningful features from EMG signals due to their ability to capture both spatial and temporal characteristics. EMG signals, characterized by their inherent noise and complex temporal patterns, benefit from DSCN's architecture, which isolates spatial feature extraction in the depthwise stage and integrates channel-wise information in the pointwise stage. This enables efficient detection of subtle muscle activity patterns and dynamic changes in the signal. DSCNs are lightweight convolutional neural network (CNN) architectures designed to minimize the number of parameters and computational complexity, making them ideal for real-time applications with limited computational resources, such as wearable devices or prosthetics. Unlike traditional CNNs, where each convolutional kernel processes all input channels, DSCNs reduce the computational load by applying each kernel to a single channel at a time, significantly lowering the number of calculations required that lead to reduce the computational cost or FLOPS. 
The DSCN architecture, illustrated in Figure \ref{fig:sep_tcn}, achieves efficiency by dividing the convolution operation into two stages:

\paragraph{Depthwise Convolution}
Depthwise convolution is particularly advantageous for EMG signal-based hand gesture recognition due to its ability to efficiently capture channel-specific spatial features from inherently noisy and complex temporal EMG signals. Unlike traditional convolutional neural networks (CNNs), which use a 3D kernel encompassing width, height, and the number of channels, depthwise convolution applies smaller, independent kernels to each input channel separately. This targeted approach reduces computational complexity while preserving each channel's unique characteristics. Mathematically, the depthwise convolution can be expressed as:
\[
Y_{\text{depth}} = X * K_{\text{depth}}
\]
where \( X \) is the input tensor, \( K_{\text{depth}} \) is the depthwise convolution kernel, and \( Y_{\text{depth}} \) represents the output feature map generated for each channel.

By isolating individual channels, depthwise convolution allows the model to extract localized spatial features and subtle variations in muscle activity, which are critical for accurate EMG signal analysis.

\paragraph{Pointwise Convolution}
Pointwise convolution complements depthwise convolution by combining the channel-specific feature maps into a comprehensive representation. Using a \( 1 \times 1 \) kernel, pointwise convolution performs channel-wise summation, enabling the integration of spatially distinct features extracted during depthwise convolution. This operation is mathematically represented as:
\[
Y = Y_{\text{depth}} * K_{\text{point}}
\]
where \( K_{\text{point}} \) is the pointwise convolution kernel, and \( Y \) is the final output tensor.

\paragraph{Depthwise Separable Convolutional Network (DSCN) Formula}
The complete Depthwise Separable Convolutional Network (DSCN) operation is given by:
\[
Y = \text{PointwiseConv}(\text{DepthwiseConv}(X, K_{\text{depth}}), K_{\text{point}})
\]
This two-step process reduces the number of parameters and computations compared to standard convolution, making it ideal for resource-constrained environments like wearable EMG systems. By focusing on localized spatial features during depthwise convolution and synthesizing global channel interactions during pointwise convolution, DSCNs provide a computationally efficient yet effective framework for extracting meaningful features from EMG signals. This combination is critical for real-time and low-latency applications such as prosthetics and gesture recognition systems, where precision and efficiency are paramount.

\paragraph{Benefits of DSCN}
By separating the convolution operation into depthwise and pointwise steps, DSCNs decouple the spatial and channel correlation. The depthwise convolution handles channel-wise correlations independently, and the pointwise convolution merges the feature maps from all channels. This separation reduces computational complexity and speeds up model training and inference. Although using DSCN results in a more significant number of network layers due to the two-step process, it significantly decreases the number of parameters in the model. This reduction leads to lower storage and computational requirements, as well as improved model efficiency. As a result, the DSCN technique will be used in this paper, replacing the traditional CNN structure in the model design.

{\begin{figure}[ht]
\centering
\includegraphics[scale=0.18]{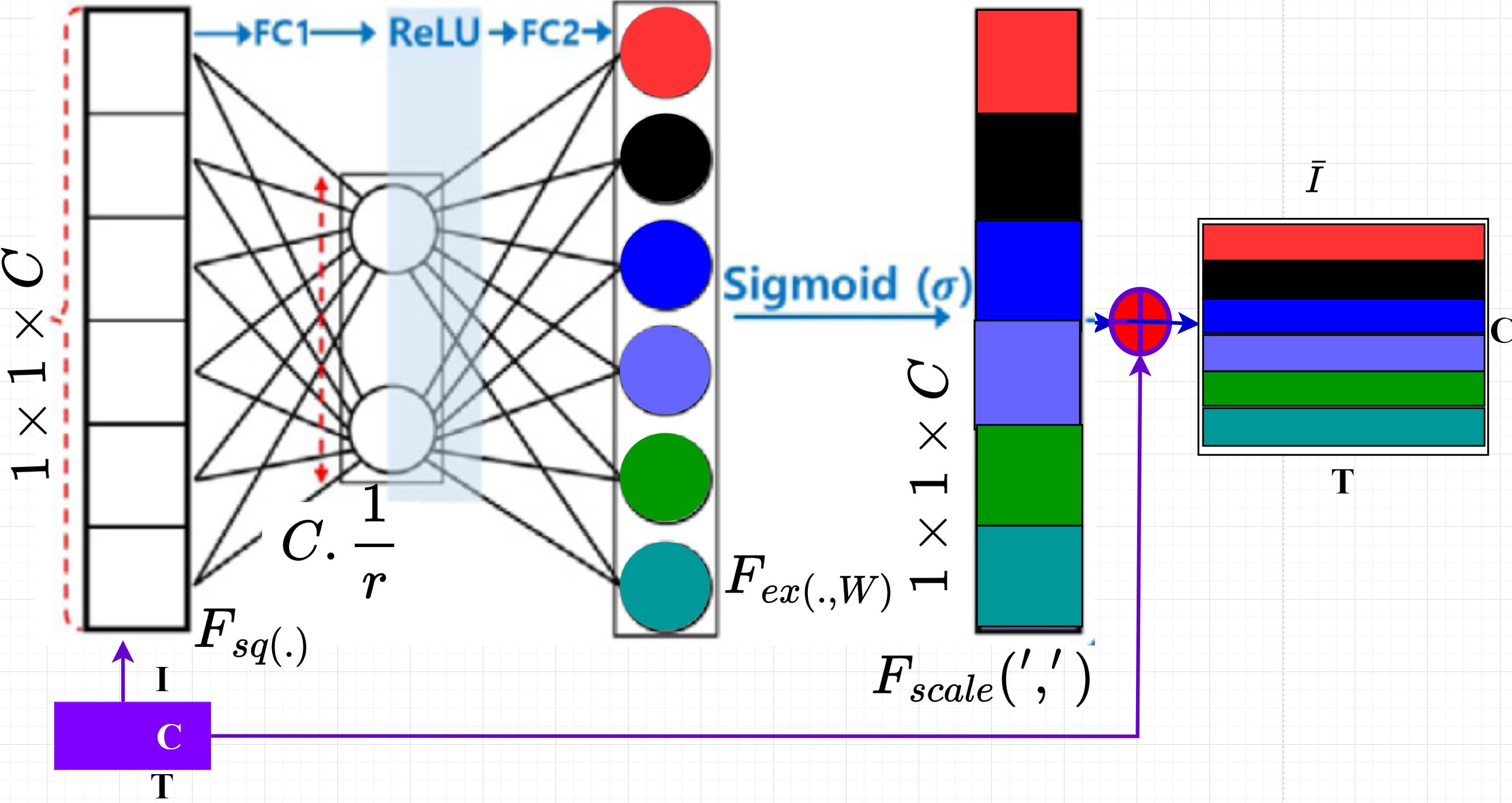}
\caption{SE block structure \cite{ma2024ffcslt_main_traffic_emg}.} %park2022advanced_se,
\label{fig:se_block}
\end{figure} }
\subsubsection{SE Block}
The Squeeze-and-Excitation (SE) block \cite{hu2018squeeze_35} enhances feature extraction from sEMG signals after the separable Temporal Convolutional Network (sep-TCN). It recalibrates the importance of each feature channel by modelling inter-channel dependencies, allowing the network to prioritize meaningful information while suppressing irrelevant noise. This is particularly advantageous for sEMG signals, which often exhibit variability due to electrode placement, muscle activity, and signal acquisition conditions. The SE block consists of two main operations: \textit{squeeze} and \textit{excitation}, as illustrated in Figure \ref{fig:se_block}. Given an input feature map \( I \in \mathbb{R}^{T \times C} \), where \( T \) is the temporal dimension and \( C \) is the number of channels, the \textit{squeeze} operation aggregates information across the temporal dimension to generate a channel descriptor. This is achieved using global average pooling:
\begin{equation}
z_c = \frac{1}{T} \sum_{t=1}^T I_{t,c}, \quad c \in \{1, \dots, C\},
\label{eq:squeeze}
\end{equation}
where \( z_c \) is the aggregated descriptor for channel \( c \).

The \textit{excitation} operation then applies two fully connected (FC) layers with non-linear activations to model channel dependencies and produce new channel weights. This is defined as:
\begin{equation}
s = \sigma(W_2 \delta(W_1 z)),
\label{eq:excitation}
\end{equation}
where \( W_1 \) and \( W_2 \) are learnable weight matrices, \( \delta \) represents the ReLU activation, and \( \sigma \) denotes the sigmoid activation.

Finally, the recalibrated feature map is obtained by rescaling the input channels:
\begin{equation}
\tilde{I}_{t,c} = s_c \cdot I_{t,c}, \quad t \in \{1, \dots, T\}, \ c \in \{1, \dots, C\}.
\label{eq:recalibration}
\end{equation}

The SE block adaptively amplifies critical channel information while suppressing noise, improving the robustness and accuracy of feature extraction. In the context of sEMG-based gesture recognition, this mechanism ensures that the model focuses on the most informative channels, often corresponding to specific muscle activities. By leveraging this selective attention, the SE block enhances the overall classification performance, particularly in scenarios involving complex or noisy sEMG signal patterns.

\paragraph{Squeeze Operation}
The squeeze operation in the SE block is a critical step that captures global information about each feature channel. This operation applies global average pooling to the input feature map \( I \) along the time dimension \( T \), effectively condensing the temporal information into a single statistic for each channel. Given an input feature map \( I = [I_1, I_2, \dots, I_C] \), where \( I_c \in \mathbb{R}^T \), the squeeze operation produces a channel descriptor \( x \in \mathbb{R}^C \). The \( c \)-th element of \( x \) is calculated as:
\begin{equation}
x_c = F_{\text{sq}}(I_c) = \frac{1}{T} \sum_{i=1}^{T} I_c(i),
\label{eq:squeeze_operation}
\end{equation}

where \( x_c \) represents the average value of the \( c \)-th channel over the time dimension \( T \).

The squeeze operation is essential for capturing global contextual information, particularly advantageous for sEMG signals. By aggregating temporal data, the squeeze operation generates a compact representation that reflects the overall activity of each channel. This reduces the influence of temporal noise or irrelevant variations, enabling the model to focus on meaningful patterns. In the context of gesture recognition, this operation ensures that the SE block can effectively identify and prioritize channels with significant contributions to the gesture classification task. By distilling the temporal information into a concise descriptor, the squeeze operation enhances the robustness and efficiency of the SE block, ultimately improving the model's accuracy and generalization capabilities.

\paragraph{Excitation Operation}
The excitation operation is the second critical step in the SE block, designed to capture channel dependencies using the channel statistics obtained from the squeeze operation. This operation leverages two fully connected layers: a dimensionality reduction layer with weight parameter \( Y_1 \) and reduction ratio \( r \), and a dimensionality expansion layer with weight parameter \( Y_2 \). In this study, the ReLU activation function is employed in place of the Sigmoid function due to its computational efficiency, ability to mitigate the vanishing gradient problem, and reduced risk of overfitting. The excitation operation is mathematically defined as:

\begin{equation}
w = F_{\text{ex}}(x, Y) = \sigma(Y_2 \cdot \sigma(Y_1 \cdot x)),
\label{eq:excitation_operation}
\end{equation}

where \( \sigma \) represents the ReLU activation function, \( Y_1 \in \mathbb{R}^{C \times (C/r)} \), \( Y_2 \in \mathbb{R}^{(C/r) \times C} \), and \( w \in \mathbb{R}^C \) is the weight vector used for channel recalibration.

\paragraph{Scale Operation}
The final step in the SE block is the scale operation, which applies the learned weights \( w \) to the input feature map \( I_c \) via channel-wise multiplication. This operation recalibrates each channel, enhancing its importance or suppressing it based on the learned weights. The scaled output \( \tilde{I} \) is computed as:

\begin{equation}
\tilde{I} = F_{\text{scale}}(I, w) = [w_1 \cdot I_1, w_2 \cdot I_2, \dots, w_C \cdot I_C],
\label{eq:scale_operation}
\end{equation}

where \( w_c \) is the learned weight for channel \( c \), and \( F_{\text{scale}}(I, w) \) represents the channel-wise multiplication between \( w_c \) and \( I_c \).

\paragraph{Advantages and Relevance}
The SE block's combination of excitation and scaling operations enables the model to selectively amplify relevant feature channels while suppressing irrelevant or noisy ones. This recalibration improves the model's focus on informative features, which is particularly beneficial for sEMG-based gesture recognition tasks. By dynamically adapting the channel importance, the SE block enhances feature discrimination and robustness, ultimately boosting classification accuracy and generalization. These characteristics make it an essential component of the proposed model, particularly in processing complex temporal and spatial dependencies in sEMG signals.

{\begin{figure*}[ht]
\centering
\includegraphics[scale=0.25]{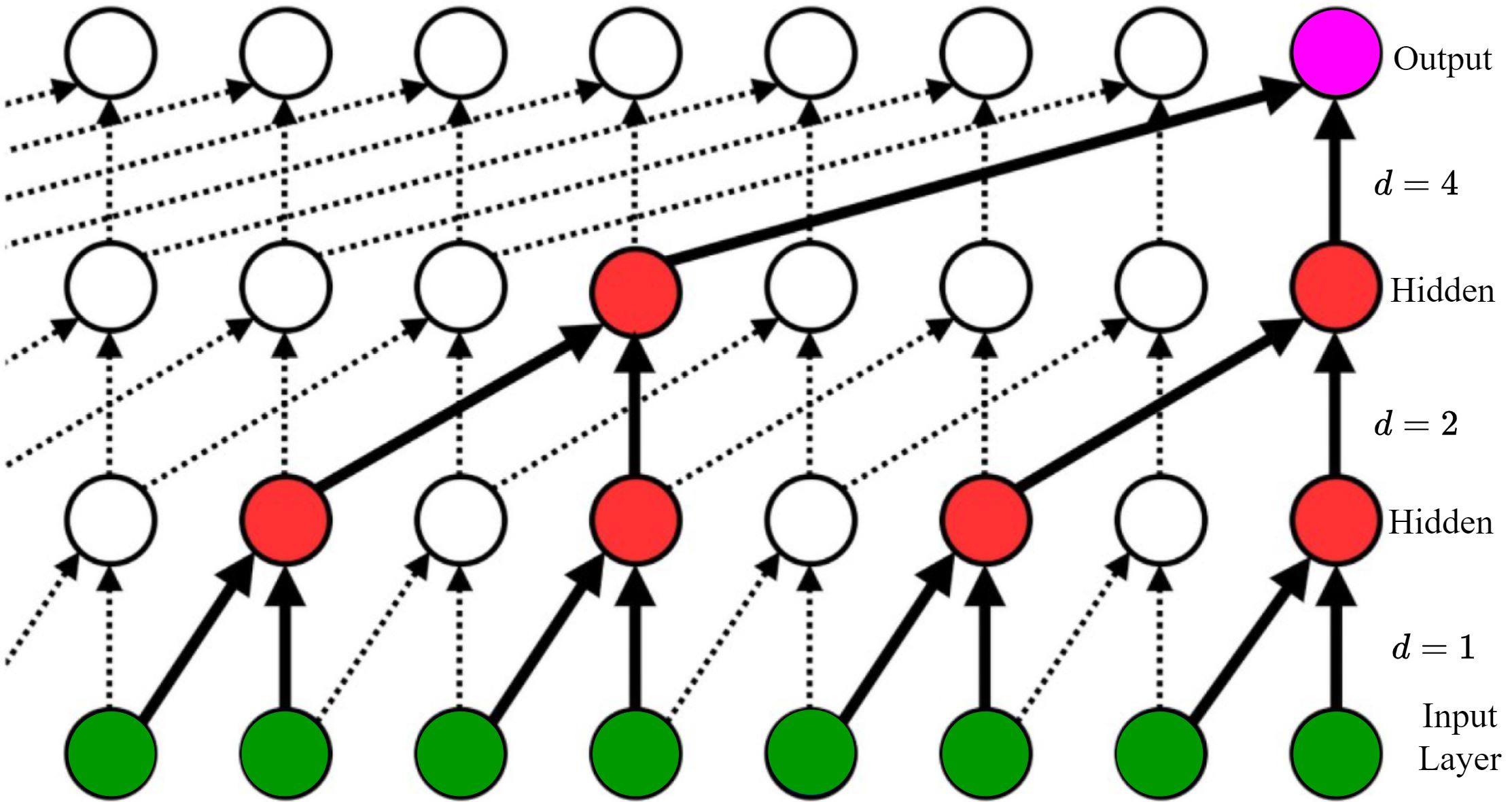}
\caption{The workflow architecture of the TCN. The layer of this module corresponds to an exponentially increasing
dilation factors d = 1, 2, 4. The green circles correspond to an input layer; the red circles correspond to hidden layers, and the cyan circles correspond to an output layer \cite{TCN_betthauser2019stable,ma2024ffcslt_main_traffic_emg}.}
\label{fig:tcn}
\end{figure*} }
\subsection{Stream-3: TCN and Bi-LSTM}
\label{meth:mult3}
The third stream utilizes Temporal Convolutional Networks (TCN) followed by BiLSTM to capture temporal dependencies and extract compelling temporal features.
Here, we fed the output of the TCN module in the Bi-LSTM module.  By combining TCN's ability to capture long-range dependencies with the bidirectional nature of LSTM, our architecture benefits from the strengths of both techniques, effectively learning from both past and future time steps while maintaining computational efficiency.
Bidirectional Long Short-Term Memory (BiLSTM) extends the standard LSTM architecture, designed to capture both past and future context in sequential data. While traditional LSTM processes data in a single direction (usually forward in time), BiLSTM processes the input sequence in both forward and backward directions. This allows the model to consider future context as well as past context when making predictions, providing a more comprehensive understanding of the sequence.

The primary advantage of BiLSTM is its ability to capture complex dependencies in time series data by processing information in both directions. This feature is handy for applications such as hand gesture recognition, where both the previous and future frames contribute valuable context to the recognition task. In our proposed system, the integration of BiLSTM with TCN enhances the model’s ability to capture intricate temporal relationships in the sEMG data, allowing for more accurate and robust hand gesture classification.

Let \( x(t) \in \mathbb{R}^d \) represent the input sequence at time step \( t \), where \( d \) is the input dimensionality (e.g., the number of EMG channels), and \( T \) is the total number of time steps. The output from the TCN branch, \( y_{\text{TCN}}(t) \), is passed as the input to the BiLSTM layer.
The TCN output \( y_{\text{TCN}}(t) \) at each time step is fed into the BiLSTM, where the BiLSTM processes the sequence in two directions — forward and backward. This is mathematically described as:
\begin{equation}
\mathbf{y}_{\text{TCN}}(t) = \text{TCN}(x(t)) \quad \text{for each time step} \ t
\label{eq:tcn_output}
\end{equation}
Let the output sequence from the BiLSTM be denoted as \( \mathbf{h}(t) \), where:
\begin{equation}
\mathbf{h}(t) = \mathbf{h}_\text{forward}(t) + \mathbf{h}_\text{backward}(t)
\label{eq:bilstm_output}
\end{equation}

Here \( \mathbf{h}_forward(t) \) is the output from the forward LSTM, which processes the sequence from time step \( t = 1 \) to \( t = T \), \( \mathbf{h}_backward(t) \) is the output from the backward LSTM, which processes the sequence from time step \( t = T \) to \( t = 1 \). The forward and backward LSTM outputs at time step \( t \) are combined, either by concatenation or addition, as shown in equation \(\ref{eq:bilstm_output}\).
For both forward and backward LSTM, the hidden state at time \( t \), \( h(t) \), is updated using the LSTM's cell state and previous hidden state. Let’s define the forward and backward LSTM's updates as follows.

The forward LSTM processes the sequence in the normal direction, and its update rule is given by:
\begin{equation}
\mathbf{h}_\text{forward}(t) = \text{LSTM}_{\text{forward}}(y_{\text{TCN}}(t), \mathbf{h}_\text{forward}(t-1), \mathbf{c}_\text{forward}(t-1))
\label{eq:forward_lstm}
\end{equation}

Where: \( \mathbf{h}_forward(t) \) is the hidden state of the forward LSTM at time step \( t \),  \( \mathbf{c}_\text{forward}(t) \) is the cell state of the forward LSTM, \( y_{\text{TCN}}(t) \) is the input to the forward LSTM (output of the TCN at time step \( t \)). 
The backward LSTM processes the sequence in reverse order, and its update rule is:
\begin{equation}
\mathbf{h}_\text{backward}(t) = \text{LSTM}_{\text{backward}}(y_{\text{TCN}}(t), \mathbf{h}_\text{backward}(t+1), \mathbf{c}_\text{backward}(t+1))
\label{eq:backward_lstm}
\end{equation}
Where:
\( \mathbf{h}_\text{backward}(t) \) is the hidden state of the backward LSTM at time step \( t \), \( \mathbf{c}_\text{backward}(t) \) is the cell state of the backward LSTM.
The output sequence \( \mathbf{h}(t) \) from the BiLSTM layer fed into the average pooling layer produces the feature of the branch-1. By feeding the TCN output into the BiLSTM layer, our architecture takes advantage of the temporal dependencies captured by the TCN, while the BiLSTM layer further enhances the model by considering both past and future context. This combination is particularly powerful for tasks like hand gesture recognition from sEMG signals, where understanding both the preceding and following frames is essential for accurate classification.
    
\subsection{Network Structure}The proposed network architecture consists of three distinct streams designed to capture both temporal and spatial features from the sEMG signals for improved hand gesture recognition. The first stream focuses on temporal learning and begins with a Temporal Convolutional Network (TCN) to capture long-range temporal dependencies, followed by a Bidirectional Long Short-Term Memory (BiLSTM) layer that captures both forward and backward temporal patterns. An average pooling layer then processes the output to reduce the temporal dimensionality while retaining key features, enhancing the model's ability to recognize dynamic gestures. The second stream is dedicated to spatial feature extraction and starts with a 1-D Convolutional (CNN) layer to capture local patterns, followed by a separable CNN to efficiently model more complex temporal dependencies. The features are then refined using a Squeeze-and-Excitation (SE) block, which recalibrates channel-wise feature maps to emphasize the most informative features and is processed by an average pooling layer to reduce dimensionality. This stream uniquely combines the separable CNN and SE block for efficient feature extraction while focusing on spatial-temporal interactions. The third stream employs a Bidirectional Temporal Convolutional Network (Bi-TCN) to capture intricate long-range temporal dependencies in both directions, followed by an average pooling operation to reduce dimensionality and retain critical temporal information. The outputs of all three streams are concatenated and passed through a flattened layer to generate low-dimensional feature vectors. These vectors are then input into a fully connected layer to extract high-level associations between the fused features, followed by a softmax layer for classification. This multi-stream architecture effectively combines various deep learning techniques—TCN, BiLSTM, separable CNN, SE block, and Bi-TCN—enabling the model to capture a comprehensive understanding of the complex dynamics of muscle activity in sEMG signals, leading to improved gesture recognition accuracy. Fusing features from multiple streams, combined with channel attention for feature selection, ensures that the model can focus on the most informative features, making the system more efficient, robust, and accurate for real-world applications.

\subsection{Feature Concatenation and Selection}
\label{meth:feat}
Features from different branches (such as standard TCN and Bi-TCN) are fused together (e.g., concatenated) to form a final feature set. A channel attention mechanism \cite{miah2024sign_largescale,miah_reivew2024methodological,egawa2023dynamic_fall_miah,10360810_miah_ksl2,shin2023korean_ksl1,shin2024korean_ksl0,shin2024japanese_jsl1,electronics12132841_miah_multistream_4} finally applied to focus on the most relevant features and reduce computational complexity while preserving key temporal and spatial features.
\subsection{Classification}
\label{meth:classification}
The fused features are passed through a fully connected layer followed by a softmax layer to classify the hand gestures.

\section{Experimental Results}
\label{res:result}
In the study, we used three benchmark datasets, including Ninapro DB2, DB4, and DB5, to evaluate the proposed model and we used Leave-One-Out Cross-Validation (LOOCV) for training and testing. In the section we described environmental setting, the performance accuracy and state of the art comparison for each datasets. 
\subsection{Environmental Setting}
 In the study, we conducted our experiment via Python programming language, where our model was trained over 100 epochs with a learning rate of 0.001 and a batch size of 32 to balance memory constraints. Preprocessed sEMG signals were directly input as multidimensional time series. To minimize prediction biases and reduce overfitting, ten-fold cross-validation (CV) was applied, dividing data into ten folds, where each served as a test set once, and the remaining nine were used for training. The averaged results of these validations provided robust performance estimates. Additionally, Leave-One-Out Cross-Validation (LOOCV) was employed to assess performance, offering an unbiased evaluation despite its computational intensity. Evaluation metrics included Accuracy, Loss, Precision, Recall, F1-score, and AUC. To handle the unique properties of sEMG signals, the model used smaller convolution kernels and filter sizes.

% \subsubsection*{NinaPro DB1 sEMG Dataset }
% \textcolor{blue}{
% The Ninapro DB1 dataset includes sEMG signals and hand gesture data from 27 subjects (20 males, 7 females; 25 right-handed, 2 left-handed) with an average age of 28.0 years (SD = 3.4 years) \cite{atzoribenchmark_DB1, atzori2014electromyography_DB1, atzori2012building_DB1}.
% This dataset captures 52 different hand movements using 10 sEMG electrodes, with each subject performing 10 repetitions per gesture.
% It includes three exercises focusing on various hand movements, from basic finger motions to complex grasping actions.
% Data entries encompass subject IDs, exercise details, sEMG and glove signals, stimulus labels, and repetition indicators.
% Table \ref{tab:dataset} shows information about the DB1 dataset.}

\subsection{Experimental Setup and Hyperparameter Tuning}

The proposed model was developed using Python 3.8 and tested with TensorFlow 2.4.1, along with pandas, NumPy, and sci-kit-learn, keras-tin, ensuring efficient data handling and computation. Implementation was conducted on a machine with an Intel Core i7-9700K processor, 16 GB RAM, and Ubuntu 20.04 LTS, providing reliable performance for training and evaluation. Similar to some previous works, the data was split into training and testing sets in a ratio of 4:2. More specifically, gesture numbers [1, 3, 4, 6]: corresponding to the 1st, 3rd, 4th, and 6th trials were used for training data, and gesture numbers [2, 5]: corresponding to the 2nd and 5th trials were used for testing data. Adam was used as the optimizer, with the learning rate set to 0.01 for DB2 and DB5 experiments and 0.0025 for DB4 experiments. Preprocessing included normalization of sEMG signals to a [0, 1] range for consistency, along with data augmentation via random cropping and flipping to improve generalization. Hyperparameters were optimized for both accuracy and efficiency. The BiLSTM layers had 32 units for capturing temporal features, while the CNN layers used 2 filters with a kernel size of 3 for spatial feature extraction. TCN and BiTCN layers were configured with 32 filters and one residual block to balance pattern recognition and efficiency. The training process utilized a batch size of 32, with 128 epochs for DB1 and 16 epochs for DB9, ensuring adaptability to dataset specifics. The selection of these hyperparameters was guided by empirical testing and domain knowledge. The learning rate was tuned based on preliminary experiments, with Adam and SGD optimizers providing a balance between convergence speed and stability. A kernel size of 3 was chosen for both Conv1D and SeparableConv1D layers, as smaller kernels effectively capture local temporal patterns in sEMG signals. The TCN was configured with 32 filters and a kernel size of 3 to capture long-range temporal dependencies, with a single stack used to prevent overfitting while maintaining sufficient representational power. To reduce overfitting, a dropout rate of 0.2 was applied after concatenation. A channel attention block was also incorporated to enhance the model's ability to focus on relevant features. These hyperparameters were selected based on a balance between model performance and computational efficiency, with potential for further optimization through advanced techniques such as grid search or Bayesian optimization in future work.

% \begin{table}[H]
% \centering
% \caption{Ablation study of the proposed model}
% \label{tab:ablation study table}
% %\setlength{\tabcolsep}{8pt}
% \begin{adjustwidth}{0cm}{0cm} % Adjust content width
% \begin{tabular}{|l|l|l|l|l|}
% \hline
% \begin{tabular}[c]{@{}l@{}}Ablation\\  Study\end{tabular} & \begin{tabular}[c]{@{}l@{}}Number\\  of \\ TCNs\end{tabular} & \begin{tabular}[c]{@{}l@{}}Skip \\ Conne-\\ction\end{tabular} & \begin{tabular}[c]{@{}l@{}} Accuracy \\ with  DB1 \\{[}\%{]}\end{tabular} & \begin{tabular}[c]{@{}l@{}}Performance \\Accuracy \\with DB9 \\{[}\%{]}\end{tabular} \\
% \hline
% Ablation-1 & 1 & - & 91.48 & 94.94 \\ \hline
% Ablation-2 & 2 Sequential & - & -3.00 & -3.00 \\ \hline
% Ablation-3 & 3 Sequential & - & -3.00 & -3.00 \\ \hline
% Ablation-4 & 2 Parallel  & - & +2.00 & -2.00 \\ \hline
% \begin{tabular}[c]{@{}l@{}}Proposed model \\without \\channel attention\end{tabular}  & 4 Stream & 1 & +2.00 & -1.00\\ \hline
% \begin{tabular}[c]{@{}l@{}}Proposed model \\with \\channel attention\end{tabular}  & 4 Stream & 1 & +3.00& +4.00\\
% \hline
% \end{tabular}
% \end{adjustwidth}
% \end{table}
\subsection{Ablation Study}
% \usepackage{multirow}
The ablation study evaluates the significance of each component in the proposed three-stream architecture for sEMG-based hand gesture recognition, shown in Table \ref{tab:ablation_study}. This architecture integrates Bidirectional-TCN, CNN, and BiTCN streams, complemented by a channel attention module, to enhance feature extraction and classification accuracy. The study examines the impact of excluding or modifying individual components on the overall model performance. For instance, removing the channel attention module reduces accuracy to 90.73\%, highlighting its critical role in dimensionality reduction and feature selection. Similarly, omitting the BiTCN, CNN, and BiLSTM branches results in significant performance drops, achieving accuracies of 64.87\%, 90.52\%, and 90.42\%, respectively. The full model, which includes all three streams and the attention mechanism, achieves the highest accuracy of 92.40\% for DB4 and 93.34\% accuracy for the DB5 datasets. This comprehensive evaluation demonstrates the robustness and effectiveness of the integrated architecture. By combining spatial and temporal feature extraction with advanced attention mechanisms, the proposed model achieves superior accuracy and computational efficiency, validating its utility in complex sEMG-based gesture recognition tasks. The findings from the ablation study highlight the synergy among the architectural components, reinforcing the contributions of the proposed design to improving recognition performance and robustness.
\begin{table*}[h]
\centering
\caption{Ablation Study Results for DB4 and DB5 Dataset}
\label{tab:ablation_study}
\begin{tabular}{|c|c|c|c|c|c|c|}
\hline
\textbf{Study} & \textbf{Branch-1 (BiLSTM)} & \textbf{Branch-2 (CNN)} & \textbf{Branch-3 (BiTCN)} & \textbf{Ch-Attention} & \textbf{DB4(\%)}& \textbf{DB5(\%)} \\ \hline
1              & Yes                         & Yes                     & Yes                      & \textbf{No}          & 90.73      &   91.63                   \\ \hline
2              & Yes                         & Yes                     & \textbf{No}             & Yes                  & 64.87   &        77.70                  \\ \hline
3              & Yes                         & \textbf{No}            & Yes                      & Yes                  & 90.52   &        91.88                 \\ \hline
4              & \textbf{No}                & Yes                     & Yes                      & Yes                  & 90.42   &         91.25                  \\ \hline
Proposed       & Yes                         & Yes                     & Yes                      & Yes                  & 92.40 &       93.34                      \\ \hline
\end{tabular}
\end{table*}

% \subsection*{Accuracy on the DB2 Dataset}
% \subsection*{Accuracy on the DB3 Dataset}
\begin{table}[h!]
\centering
\caption{Performance Metrics of the DB2 Dataset Across Subjects}
\label{tab:acc_db2_performance_metrics}
\begin{tabular}{|p{1cm}|p{1cm}|p{1cm}|p{1cm}|p{1cm}|}
\hline
\textbf{Subject} & \textbf{Accuracy (\%)} & \textbf{Precision (\%)} & \textbf{Recall (\%)} & \textbf{F1-Score (\%)} \\ \hline
1 & 96.09 & 98.96 & 98.44 & 98.33 \\ \hline
2 & 99.41 & 98.96 & 98.44 & 98.33 \\ \hline
3 & 96.02 & 95.83 & 93.75 & 93.33 \\ \hline
4 & 95.23 & 98.96 & 98.44 & 98.33 \\ \hline
5 & 95.59 & 98.96 & 98.44 & 98.33 \\ \hline
6  & 95.27 & 95.96 & 93.94 & 94.34 \\ \hline
7  & 95.78 & 97.92 & 96.88 & 96.67 \\ \hline
8  & 95.66 & 94.95 & 92.42 & 92.73 \\ \hline
9  & 96.64 & 95.96 & 93.94 & 94.34 \\ \hline
10 & 94.65 & 95.96 & 93.94 & 94.34 \\ \hline
11 & 94.77 & 97.92 & 96.88 & 96.67 \\ \hline
12 & 96.21 & 94.95 & 92.42 & 92.73 \\ \hline
13 & 96.25 & 98.96 & 98.44 & 98.33 \\ \hline
14 & 95.70 & 97.92 & 96.88 & 96.67 \\ \hline
15 & 94.34 & 97.92 & 96.88 & 96.67 \\ \hline
16 & 96.76 & 94.95 & 92.42 & 92.73 \\ \hline
17 & 95.51 & 96.35 & 95.31 & 95.10 \\ \hline
18 & 94.61 & 98.96 & 98.44 & 98.33 \\ \hline
19 & 95.86 & 97.92 & 96.88 & 96.67 \\ \hline
20 & 96.09 & 95.96 & 93.94 & 94.34 \\ \hline
21 & 94.22 & 98.96 & 98.44 & 98.33 \\ \hline
22 & 95.66 & 94.95 & 92.42 & 92.73 \\ \hline
23 & 96.41 & 97.92 & 96.88 & 96.67 \\ \hline
24 & 95.08 & 97.92 & 96.88 & 96.67 \\ \hline
25 & 94.77 & 97.92 & 96.88 & 96.67 \\ \hline
26 & 96.68 & 96.88 & 95.31 & 95.00 \\ \hline
27 & 94.77 & 98.96 & 98.44 & 98.33 \\ \hline
28 & 96.09 & 98.96 & 98.44 & 98.33 \\ \hline
29 & 95.15 & 97.92 & 96.88 & 96.67 \\ \hline
30 & 96.05 & 98.96 & 98.44 & 98.33 \\ \hline
31 & 90.98 & 94.95 & 92.42 & 92.73 \\ \hline
32 & 90.74 & 96.88 & 95.31 & 95.00 \\ \hline
33 & 96.41 & 98.96 & 98.44 & 98.33 \\ \hline
34 & 96.33 & 98.96 & 98.44 & 98.33 \\ \hline
35 & 96.91 & 93.94 & 90.91 & 91.11 \\ \hline
36 & 95.39 & 95.96 & 93.94 & 94.34 \\ \hline
37 & 91.91 & 97.92 & 96.88 & 96.67 \\ \hline
38 & 92.93 & 97.92 & 96.88 & 96.67 \\ \hline
39 & 92.89 & 88.89 & 86.36 & 85.66 \\ \hline
40 & 96.41 & 98.96 & 98.44 & 98.33 \\ \hline
Average  & 95.31 & - & - & - \\ \hline
\end{tabular}
\end{table}
\subsection{Peformance Accuracy on the DB2 Result}
The resulting Table \ref{tab:acc_db2_performance_metrics} comprehensively evaluates the model's performance on various subjects, using key metrics including accuracy, precision, recall, and F1-score. The table highlights the model's consistent ability to achieve high accuracy, with most subjects scoring from subjects 1-5 above 95\%, underscoring its robustness and reliability across diverse data inputs. Notably, Subject 2 achieves the highest accuracy of 99.41\%, accompanied by precision, recall, and F1-scores of 98.96\%, 98.44\%, and 98.33\%, respectively, demonstrating the model's exceptional performance in identifying subtle sEMG signal patterns. The table demonstrates that the model consistently achieves high accuracy, with most subjects scoring above 95\%, showcasing its robustness across diverse data inputs. Notably, Subject 13 and Subject 40 achieve the highest accuracy of 96.41\%, complemented by strong precision, recall, and F1-scores of 98.96\%, 98.44\%, and 98.33\%, respectively. This highlights the model's ability to balance accurate predictions with minimal misclassification. For subjects such as 19 and 25, the accuracy remains competitive at 95.86\% and 94.77\%, respectively, while maintaining commendable precision and recall metrics, reflecting the model's reliability in capturing intricate patterns in sEMG signals. However, Subject 39 shows a comparatively lower accuracy of 92.89\%, attributed to a precision of 88.89\% and an F1-score of 85.66\%, indicating challenges in certain data variations. However, Subject 39 shows a slightly lower accuracy of 92.89\%, with corresponding precision and F1-scores indicating challenges with data variability. Overall, the results validate the proposed architecture's effectiveness, emphasizing its suitability for sEMG-based hand gesture recognition tasks, including prosthetics and rehabilitation applications, where precision and reliability are critical. The results validate the efficacy of integrating advanced techniques like TCN, LSTM, and CNN in the proposed architecture. Consistent metrics across subjects emphasize the model's ability to generalize effectively. These findings underline the model's potential for practical applications in sEMG-based systems, such as prosthetics and rehabilitation, where accurate and reliable hand gesture recognition is critical.

% \begin{table}[]
% \caption{Performance Result with NinaPro DB2 dataset} \label{tab:accuracy_db2}
% \begin{tabular}{lllll}
% \hline
%  & Precision & Recall & F1-score & Validation Accuracy &Test Accuracy\\\hline
% Average &98.96 & 98.44 &98.33 & 98.43&96.40 \\ \hline
% \end{tabular}
% \end{table}

\begin{table}[]
\centering
\caption{Performance Result with NinaPro DB4 dataset} \label{tab:acc_db4}
\begin{tabular}{lllll}
\hline
Subject Name & Accuracy & Precision & Recall & F1-score \\\hline
1 & 94.79 & 95.14 & 94.79 & 94.44 \\ \hline
2 & 97.92 & 98.61 & 97.92 & 97.78 \\ \hline
3 & 88.54 & 88.33 & 85.00 & 85.40 \\ \hline
4 & 92.71 & 94.79 & 92.71 & 92.29 \\ \hline
5 & 90.62 & 92.71 & 90.62 & 90.21 \\ \hline
6 & 95.83 & 96.88 & 95.83 & 95.62  \\ \hline
7 & 92.71 & 93.06 & 92.71 & 91.81 \\ \hline
8 & 88.54 & 90.76 & 88.54 & 87.51 \\ \hline
9 & 89.58 & 90.97 & 89.58 & 88.40 \\ \hline
10 & 92.71 & 92.71 & 92.71 & 91.60 \\ \hline
Average & 92.40 & 94.52 & 92.96 & 92.47 \\ \hline
\end{tabular}
\end{table}
\subsection{Performance Accuracy on the DB4 Dataset}
The performance results with the NinaPro DB4 dataset, summarized in Table~\ref{tab:acc_db4}, highlight the effectiveness and adaptability of the proposed model in sEMG-based hand gesture recognition. The model achieved an average accuracy of 92.40\%, with precision, recall, and F1-scores of 94.52\%, 92.96\%, and 92.47\%, respectively. These metrics demonstrate the model's robustness across diverse subjects, as well as its ability to maintain high classification performance. The highest accuracy recorded was 97.92\%, showcasing the model's exceptional capability to identify complex gestures accurately. The combination of advanced architectural components, such as bidirectional-TCN, CNN-TCN, and channel attention modules, enables the extraction of both spatial and temporal features, significantly improving the recognition performance. The strong performance, even in cases with lower individual scores, highlights the model's resilience to challenges such as variability in muscle signals and individual subject differences. This adaptability makes the proposed system a promising solution for real-world applications, such as prosthetics and rehabilitation, where precision and reliability are critical. Furthermore, the consistent results underline the advantages of the model's multi-stream architecture and feature fusion strategy.
\begin{table}[]
\centering
\caption{Performance Result with NinaPro DB5 dataset} \label{tab:acc_db5}
\begin{tabular}{lllll}
\hline
Subject Name & DB5 Accuracy & Precision & Recall & F1-Score \\ \hline
1 & 94.79 & 96.53 & 94.79 & 94.44 \\  \hline
2 & 95.31 & 96.88 & 95.31 & 95.00 \\      \hline
3 & 97.92 & 98.61 & 97.92 & 97.78 \\   \hline
4 & 91.68 & 90.48 & 89.80 & 89.12 \\    \hline
5 & 96.88 & 96.18 & 96.88 & 96.11 \\    \hline
6 & 89.06 & 89.58 & 89.06 & 87.29 \\    \hline
7 & 89.58 & 92.71 & 89.58 & 89.44 \\    \hline
8 & 94.79 & 96.53 & 94.79 & 94.72 \\    \hline
9 & 90.63 & 92.52 & 88.78 & 88.57 \\     \hline
10 & 92.71 & 95.14 & 92.71 & 92.22 \\    \hline
Average & 93.34 & 93.40 & 92.04 & 91.51  \\ \hline
\end{tabular}
\end{table}
\subsection{Performance Accuracy on the DB5 Dataset}
The results presented in Table~\ref{tab:acc_db5} showcase the performance of the proposed model using the NinaPro DB5 dataset. The model demonstrates robust accuracy, with an average accuracy of 91.51\%, precision of 93.40\%, recall of 92.04\%, and F1-score of 97.78\%. These metrics highlight the system's strong capability in classifying sEMG-based hand gestures across diverse subjects. Subject 3 achieved the highest performance, with an accuracy of 97.92\%, precision of 98.61\%, and F1-score of 97.87\%, reflecting the model's ability to effectively capture intricate gesture patterns. On the other hand, Subject 6 reported the lowest accuracy of 89.06\%, highlighting potential variability in signal quality or subject-specific factors. The strong performance across subjects, particularly the high average precision, demonstrates the model's ability to reduce misclassifications and enhance the reliability of gesture recognition. The integration of temporal and spatial features through advanced architectures like TCN and CNN-LSTM ensures robustness against signal variability. These results affirm the proposed model's suitability for real-world applications, such as prosthetic control and rehabilitation, where high accuracy and generalizability are crucial.

\begin{table}[ht]
\centering
\caption{State of the Art Comparison for DB2 and DB4 Dataset.}
\label{tab:sota_DB2_DB4}
\begin{tabular}{|l|c|c|}
\hline
\textbf{Methods}                                    & \textbf{DB2}          & \textbf{DB4}          \\ \hline
LS-SVM (IAV+MAV+RMS+WL) \cite{nazemi2014artificial} & -                     & -                    \\ \hline
LDA (IAV(or MAV)+CC) \cite{nazemi2014artificial}    & -                     & -                    \\ \hline
RF \cite{atzori2014electromyography,pizzolato2017comparison_DB4_DB5} & 75.27\%              & 69.13 ± 7.77\%       \\ \hline
AtzoriNet \cite{atzori2014electromyography}         & 60.3 ± 7.7\%          & -                    \\ \hline
ZhaiNet \cite{zhai2017self}                         & 78.71\%               & -                    \\ \hline
TL-MKCNN \cite{zou2021transfer}                     & 86.67\%               & 82.29\%              \\ \hline
FS-HGR \cite{rahimian2021fs}                        & 85.94\%               & -                    \\ \hline
RNN with weight loss \cite{koch2018rnn}             & 78.0\%                & -                    \\ \hline

CNN \cite{geng2016gesture}            & 77.80               & -                    \\ \hline
CNN \cite{zhai2017self}               & 78.71               & -                    \\ \hline
CNN \cite{ding2018semg}               & 78.86               & -                    \\ \hline
Hybrid CNN-RNN \cite{hu2018attention} & 82.20               & -                    \\ \hline
CNN \cite{wei2019multiview}           & 83.70               & -                    \\ \hline

LSTM+MLP \cite{he2018surface}                       & -                     & -                    \\ \hline
CNN-LSTM \cite{huang2019surface}                    & 79.32\%               & -                    \\ \hline
Attention-based hybrid CNN-RNN \cite{hu2018attention} & 82.2\%               & -                    \\ \hline

Proposed Model                                      & 96.41\%               & 92.40\%              \\ \hline
\end{tabular}
\end{table}

%RCNN  \cite{xu2022novel}                                              & 87.37 ± 3.77\%        & 99.32 ± 0.55\%       & 99.25 ± 0.13\%       \\ \hline
%CFF-RCNN \cite{xu2022novel}                                           & 88.87 ± 3.63\%        & 99.51 ± 0.12\%       & 99.29 ± 0.10\%       \\ \hline
\subsection{State of the Art Comparison of the DB2 and DB4 Dataset}
The state-of-the-art comparison Table \ref{tab:sota_DB2_DB4} provides an in-depth evaluation of various methods applied to the Ninapro DB2 and DB4 datasets for sEMG-based hand gesture recognition. %Traditional machine learning approaches like LS-SVM and LDA achieve moderate accuracies of 85.19\% and 84.23\% on DB1, 
%respectively, but lack evaluations on DB2 and DB4. 
Random Forest (RF) exhibits consistent performance with %accuracies of 75.32\% on DB1,
75.27\% on DB2, and 69.13\% on DB4, indicating its limitations in handling complex datasets.
Deep learning-based models significantly improve accuracy compared to traditional methods. AtzoriNet and ZhaiNet demonstrate progress, while TL-MKCNN achieves notable performance with 86.67\% on DB2 and 82.29\% on DB4. Similarly, FS-HGR and CNN-LSTM show competitive results, with CNN-LSTM attaining 79.32\% on DB2. CNN architectures evaluated by Geng, Zhai, and Ding achieved accuracies ranging from 77.80\% to 78.86\% on DB2. Advanced methods like Hybrid CNN-RNN \cite{hu2018attention}, and CNN by Wei achieve higher accuracies of 82.20\% and 83.70\%, respectively, showcasing the potential of hybrid and multi-view learning architectures. The proposed model surpasses all existing approaches,  96.41\% on DB2 and 91.91\% on DB4. These results highlight the proposed model's superior generalization ability, robustness, and accuracy across diverse datasets, establishing it as the most effective solution for sEMG-based hand gesture recognition.

\begin{table}[ht]
\centering
\caption{Performance comparison of the proposed technique with the state-of-the-art methods on the DB4 and DB5 dataset.}
\label{tab:sota_db5_comparison}
\begin{tabular}{|p{2cm}|p{2cm}|p{1.5cm}|p{1.5cm}|}
\hline
\textbf{Author}    &\textbf{Methods}                                & \textbf{DB4 (53 Classes) Accuracy (\%)}&\textbf{DB5 (53 Classes) Accuracy (\%)} \\ \hline
Peng et all.  \cite{peng2022ensemble} & Ensemble extreme learning machine (EELM) & 77.9  &-                \\ \hline
Chen et al.  \cite{chen2023spatial}  & Decision tree (DT)      & 53.6         &-         \\ \hline
Cote et al. \cite{cote2019deep}  & DNN \cite{cote2019deep}                         & 64.65  &-               \\ \hline
Chen et al. \cite{chen2023spatial} &SVM                       & 71.25     &-            \\ \hline
Chen et al. \cite{chen2023spatial} &KNN                     & 75.41      &-           \\ \hline
Chen et al. \cite{chen2023spatial} &CNN-LSTML                       & 61.21   &-              \\ \hline
Chen et al. \cite{chen2023spatial} & LCNN                      & 66.38     &-           \\ \hline
Cote et al. \cite{cote2019deep}  & CWT-based ConvNets                       & 68.98    &-            \\ \hline
Chen et al. \cite{chen2020hand} & CWT-EMGNet \cite{chen2020hand}                      & 69.62  &-              \\ \hline
Mohapatra et al.  \cite{mohapatra2024hand}& SVM \cite{mohapatra2024hand}                      & 84.00  &-               \\ \hline
%new
Nguyen et al. (2024) \cite{nguyen} &FANet                      & 78.7 (53)     &89.6(53)            \\ \hline
Niu et al. (2024 \cite{niu} &PCS-EMGNet                     &  83.00 (53)     &88.3(53)           \\ \hline
Salerno et al. (2024) \cite{salerno} &HDC                       &  74.6 (52)     &85.7(52)              \\ \hline
Too et al. (2019) \cite{too} & SVM+LDA                     &  91.3 (40)     &-          \\ \hline
Zhou et al. (2019) \cite{zhou}  & RF                       &  -     &84.1(11)           \\ \hline
Li et al. (2023) \cite{li}  & SVM                      &  -     &90.8 (52)              \\ \hline
 Chaiyaroj et al. (2019) \cite{chaiyaroj}& DNN                     & -     &91.0(41)              \\ \hline

Proposed Model &  3-Branch    DNN           &92.4 (52)& 93.34 (52)                 \\ \hline
\end{tabular}
\end{table}

\subsection{State of the Art Comparison with DB4 and DB5 Dataset}
The performance comparison table (\ref{tab:sota_db5_comparison}) evaluates the proposed model against several state-of-the-art methods applied to the DB5 dataset for sEMG-based hand gesture recognition. Traditional machine learning approaches such as Decision Tree (DT), SVM, and KNN exhibit varying levels of accuracy, with DT achieving the lowest performance at 53.6\% and KNN slightly better at 75.41\%. SVM outperforms these models with an accuracy of 84.00\%, showcasing its ability to handle complex feature relationships effectively. Deep learning methods such as DNN and LCNN demonstrate modest improvements, achieving accuracies of 64.65\% and 66.38\%, respectively, highlighting the limitations of these architectures in capturing sEMG data's temporal and spatial intricacies. CWT-based ConvNets and CWT-EMGNet slightly improve performance, achieving accuracies of 68.98\% and 69.62\%, respectively, by leveraging time-frequency domain features. CNN-LSTML, which combines convolutional and sequential learning, achieves 61.21\%, indicating potential limitations in integrating spatial and temporal dependencies. 
The proposed model outperforms all other approaches, achieving an impressive accuracy of 93.00\%. This significant improvement underscores its ability to effectively integrate temporal and spatial features, making it the most robust and accurate solution for sEMG-based hand gesture recognition on the DB5 dataset. The results highlight the superior generalization capability and robustness of the proposed approach.

\subsection{Discussion and Practical Application}
We proposed a novel sEMG-based hand gesture recognition system that addresses challenges like unstable predictions and ineffective time-varying feature extraction in muscle-computer interfaces. The system employs a multi-branch deep learning architecture, with each branch extracting distinct spatial-temporal features to capture gesture dynamics. The first branch uses a Bidirectional Temporal Convolutional Network (Bi-TCN) to capture long-range temporal dependencies, the second integrates 1D Convolutional layers, separable CNN, and a Squeeze-and-Excitation (SE) block to emphasize critical features, and the third combines a TCN and Bidirectional LSTM (BiLSTM) to model dynamic temporal relationships. The outputs from all branches are concatenated and refined using a channel attention mechanism to emphasize the most relevant features. Evaluated on the Ninapro DB2, DB4, and DB5 datasets, the model achieved accuracies of 96.41\%, 92.4\%, and 93.34\%, respectively, outperforming traditional models (e.g., Random Forest, SVM) and recent deep learning approaches (e.g., CNN-LSTM, FANet, PCS-EMGNet), as shown in Tables \ref{tab:sota_DB2_DB4} and \ref{tab:sota_db5_comparison}.
In terms of computational complexity, the model requires 23.442 MFLOPs for an input length of 1,000 (DB5) and 234.42 MFLOPs for 10,000 (DB4), indicating its potential for real-time applications. The high accuracy and low complexity make it suitable for human-machine interface (HMI) applications, such as (1) Prosthetic Control: Decoding complex sEMG signals for intuitive prosthetic limb operation, improving the quality of life for amputees. (2) Rehabilitation Devices: Integration into assistive exoskeletons for feedback-driven therapy, aiding stroke or neuromuscular disorder patients. (3) Gesture-Based Interface Systems: Enabling natural hand gesture control in gaming, robotics, and virtual/augmented reality, enhancing user experience.
While the proposed system demonstrates strong performance, real-world applications face several challenges. One challenge is hardware variability; the sensors used in our experiments may not represent the diversity of devices in practical settings. Future work should explore low-cost, wearable sensors to ensure scalability and robustness across different hardware configurations.
Latency is another concern for real-time applications like prosthetic control and rehabilitation. Despite promising computational complexity, further optimization is required for near-instantaneous processing, particularly in environments with limited computing power. Edge computing and hardware acceleration could help minimize latency. Electrode placement variability and environmental factors such as movement, sweating, and noise can degrade sEMG signal quality in real-world conditions. To mitigate this, robust signal preprocessing techniques (e.g., noise filtering and normalization) and machine learning models capable of adapting to electrode displacement or signal distortion should be considered. Additionally, improving the model’s resilience to noise, especially during dynamic activities, will enhance performance in environments like gaming or virtual reality.

\section{Conclusions and Future Directions} 
\label{sec6}
In this study, we presented a novel sEMG-based hand gesture recognition system that effectively addresses key challenges in muscle-computer interfaces, including unstable predictions and inefficient feature enhancement. Our proposed system employs a multi-branch deep learning architecture, with each branch tailored to extract complementary features from sEMG signals, ensuring a comprehensive understanding of gesture dynamics. The first branch leverages a Bidirectional Temporal Convolutional Network (Bi-TCN) to capture long-range temporal dependencies, enabling the model to analyze past and future muscle activity for precise gesture recognition. The second branch combines 1D Convolutional layers, separable CNN, and a Squeeze-and-Excitation (SE) block to extract and emphasize critical spatial-temporal features, enhancing the system's ability to focus on informative signal components. The third branch integrates a Temporal Convolutional Network (TCN) and Bidirectional Long Short-Term Memory (BiLSTM) network, capturing both time-varying and bidirectional temporal relationships, which are crucial for dynamic gesture recognition. To ensure robust feature fusion, outputs from the three branches are concatenated and refined using a channel attention mechanism, which selectively emphasizes the most relevant features while reducing dimensionality. This innovative feature fusion approach leverages complementary information from diverse extraction methods, enhancing the model's generalization and robustness across various hand gestures. The proposed model was rigorously evaluated on the Ninapro DB2, DB4, and DB5 datasets, achieving impressive accuracy rates, thereby demonstrating its effectiveness in handling the complexities of sEMG signals and outperforming existing methods. This multi-branch architecture, with its advanced feature extraction, fusion techniques, and attention mechanisms, represents a significant step forward in gesture recognition systems. Our findings hold considerable potential for improving prosthetic limb control, enabling more intuitive and precise movements. Furthermore, this study contributes to advancements in human-machine interfaces, particularly in assistive technologies for individuals with disabilities. By enhancing the recognition of hand gestures from sEMG signals, our work paves the way for developing more efficient, reliable, and practical muscle-computer interfaces, fostering innovation and accessibility in this critical field.

%\section{Author contributions statement}
%Conceptualization, Abu Saleh Musa Miah, Jungpil Shin, Sota Konnai; Methodology, Abu Saleh Musa Miah, Jungpil Shin, Sota Konnai; Investigation, Abu Saleh Musa Miah, Sota Konnai; Data Curation, Abu Saleh Musa Miah, Sota Konnai; Writing—Original Draft Preparation, Abu Saleh Musa Miah, Sota Konnai; Writing—Review and Editing, Abu Saleh Musa Miah; Visualization, Abu Saleh Musa Miah, Sota Konnai; Supervision,Jungpil Shin; Funding Acquisition,Jungpil Shin  All authors have read and agreed to the published version of the manuscript.
% \section{Data Availability}
% \textbf{NinaPro DB2 sEMG Dataset: }%Original Source:%} 
% \url{https://ninapro.hevs.ch/instructions/DB2.html} \\
% % \textbf{NinaPro DB4 sEMG Dataset Kaggle Source:}
% % \\ \url{https://www.kaggle.com/datasets/mansibmursalin/ninapro-db1-full-dataset}\\
% \textbf{NinaPro DB4 sEMG Dataset:} \url{https://ninapro.hevs.ch/instructions/DB4.html} \\
% \textbf{NinaPro DB5 sEMG Dataset:} \url{https://ninapro.hevs.ch/instructions/DB5.html} \\

% \newpage
\bibliographystyle{unsrt}
\bibliography{reference}% common bib file
\begin{IEEEbiography}[{\includegraphics[width=1in,height=1.25in, clip,keepaspectratio]{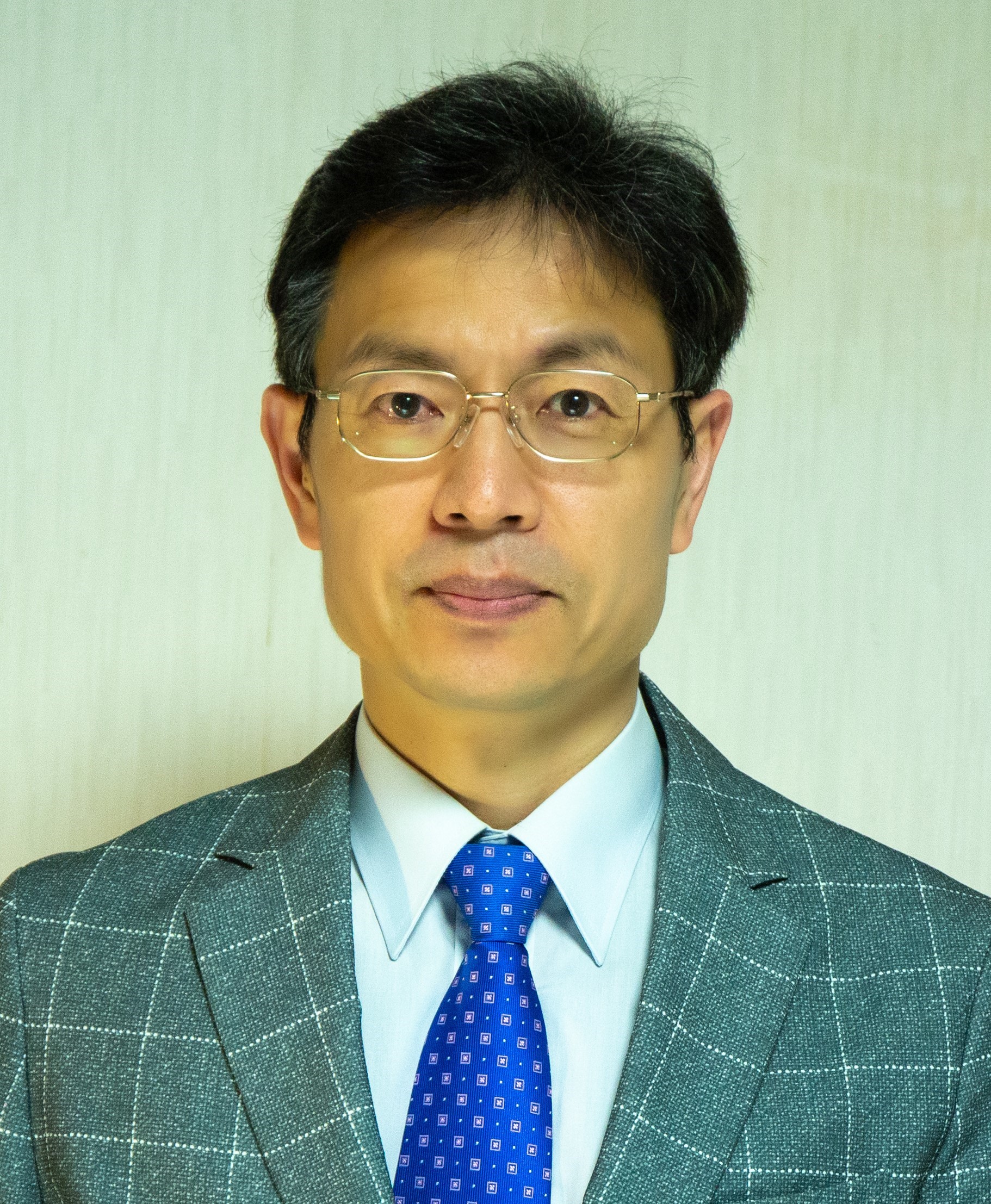}}]
{Jungpil Shin} (Senior Member, IEEE) received a B.Sc. in Computer Science and Statistics and an M.Sc. in Computer Science from Pusan National University, Korea, in 1990 and 1994, respectively. He received his Ph.D. in computer science and communication engineering from Kyushu University, Japan, in 1999, under a scholarship from the Japanese government (MEXT). He was an Associate Professor, a Senior Associate Professor, and a Full Professor at the School of Computer Science and Engineering, The University of Aizu, Japan in 1999, 2004, and 2019, respectively. His research interests include pattern recognition, image processing, computer vision, machine learning, human-computer interaction, non-touch interfaces, human gesture recognition, automatic control, Parkinson’s disease diagnosis, ADHD diagnosis, user authentication, machine intelligence, bioinformatics, as well as handwriting analysis, recognition, and synthesis. He is a member of ACM, IEICE, IPSJ, KISS, and KIPS. He served as program chair and as a program committee member for numerous international conferences. He serves as an Editor of IEEE journals Springer, Sage, Taylor and Francis, MDPI Sensors and Electronics, and Tech Science. He serves as an Editorial Board Member of Scientific Reports. He serves as a reviewer for several major IEEE and SCI journals. He has co-authored more than 350 published papers for widely cited journals and conferences.
\end{IEEEbiography}
\begin{IEEEbiography}[{\includegraphics[width=1in,height=1.25in,clip,keepaspectratio]{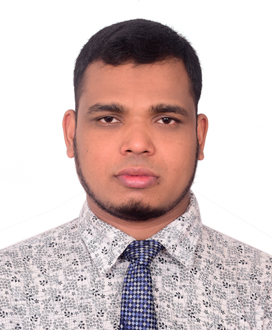}}]{Abu Saleh Musa Miah} received the B.Sc.Engg., and M.Sc.Engg. degrees in computer science and engineering from the Department of Computer Science and Engineering, University of Rajshahi, Rajshahi-6205, Bangladesh, in 2014 and 2015, respectively. He became a Lecturer and an Assistant Professor at the Department of Computer Science and Engineering, Bangladesh Army University of Science and Technology (BAUST), Saidpur, Bangladesh, in 2018 and 2021, respectively. He started his PhD degree in the School of Computer Science and Engineering at the University of Aizu, Japan, in 2021, under a scholarship from the Japanese government (MEXT). His research interests include CS, ML, DL, HCI, BCI and neurological disorder detection. He has authored and co-authored more than 20 publications published in widely cited journals and conferences.
\end{IEEEbiography}

\begin{IEEEbiography}
[{\includegraphics[width=1in,height=1.25in,clip,keepaspectratio]{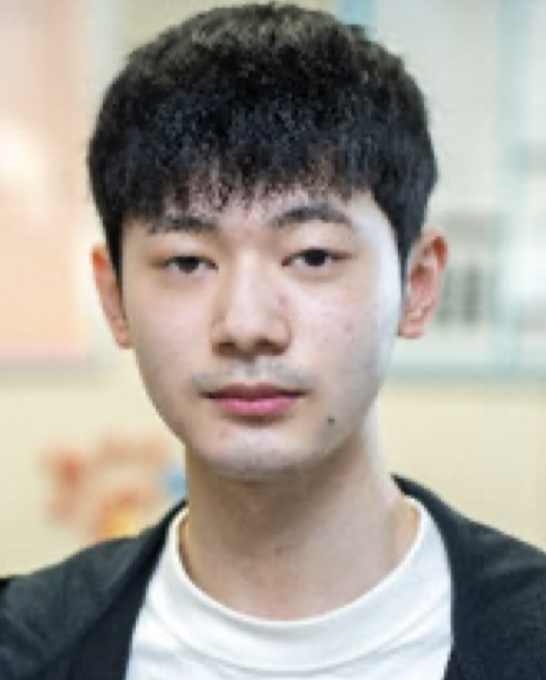}}]{SOTA KONNAI} received a bachelor’s degree in computer science and engineering from The University of Aizu (UoA), Japan, in March 2023. He is currently pursuing the master’s degree. He joined the Pattern Processing Laboratory, UoA, in April 2022, under the supervision of Prof. Dr. Jungpil Shin.  His research interests include computer vision, pattern recognition, and deep learning. He is also working on the analysis and recognition of ADHD.
\end{IEEEbiography}

\begin{IEEEbiography}
[{\includegraphics[width=1in,height=1.25in,clip,keepaspectratio]{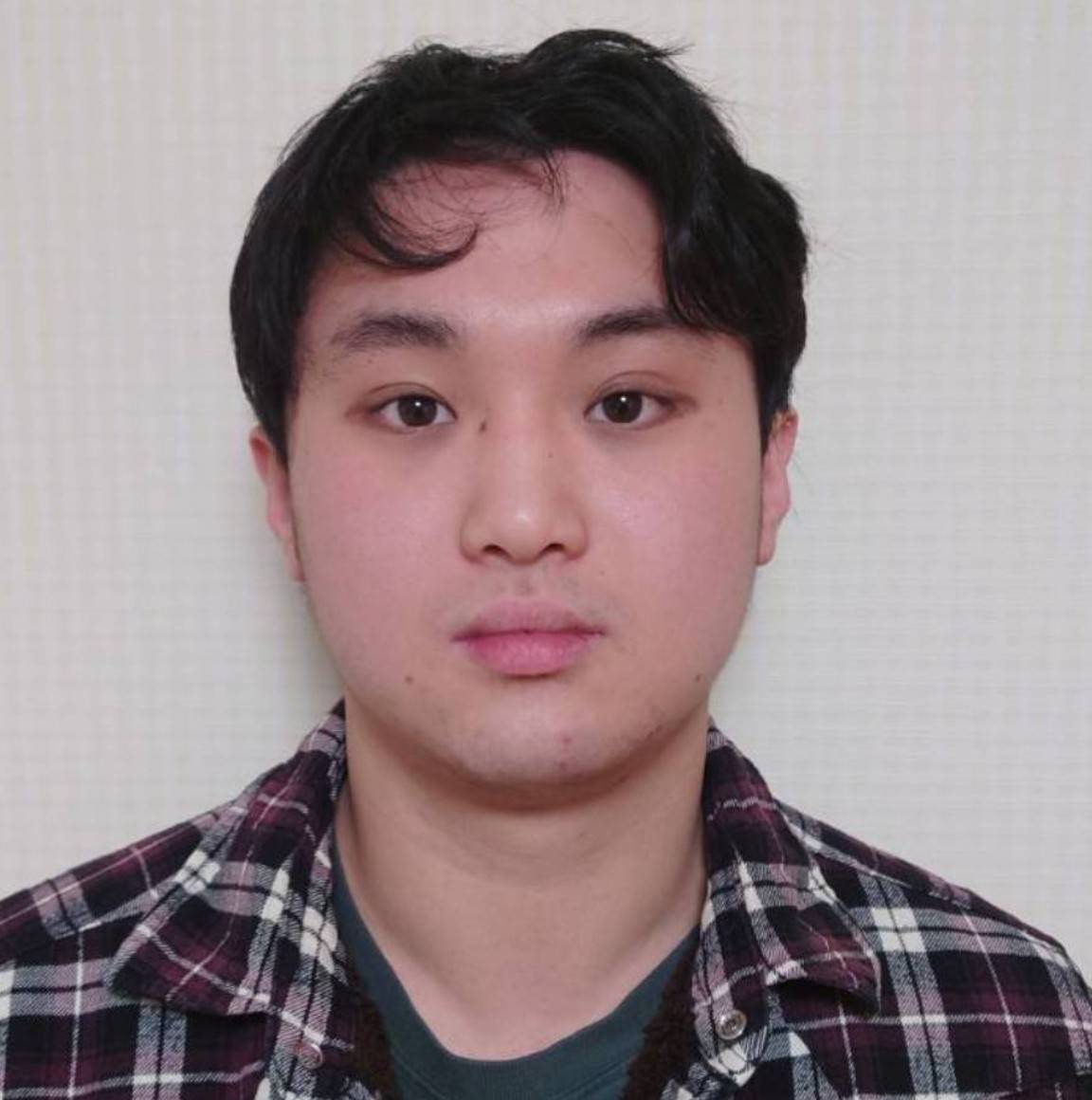}}]{HOSHITAKA SHU} is currently enrolled in the bachelor’s program in computer science and engineering at The University of Aizu (UoA), Japan. He joined the Pattern Processing Laboratory, UoA, in April 2024, under the supervision of Prof. Dr. Jungpil Shin.  His research interests include pattern recognition and deep learning.
\end{IEEEbiography}

\begin{IEEEbiography}
[{\includegraphics[width=1in,height=1.25in,clip,keepaspectratio]{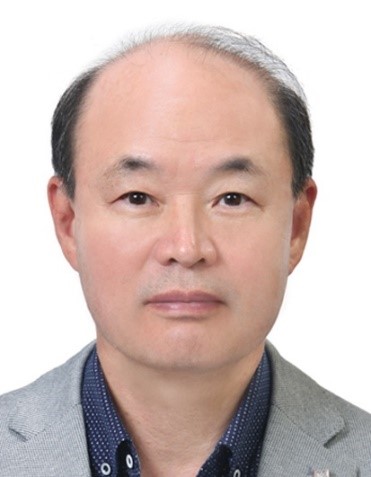}}]{PANKOO KIM} received the B.E. degree from Chosun University, in 1988, and the M.S. and
Ph.D. degrees in computer engineering from Seoul National University, in 1990 and 1994, respectively. He is currently a Full Professor with Chosun University. His research interests include semantic web techniques, semantic information processing and retrieval, multimedia processing, semantic web, and system security. He is also the Editor-in-Chief of the IT CoNvergence PRActice(INPRA) journal

\end{IEEEbiography}

\EOD
\end{document}